\documentclass[runningheads]{llncs}
\usepackage[T1]{fontenc}
\usepackage{graphicx}
\usepackage{enumitem}
\setlist[itemize]{label=$\bullet$} 

\usepackage{subcaption}  
\usepackage{multicol}  
\usepackage{longtable}
\usepackage{booktabs}
\usepackage{multicol}
\usepackage{array} 
\usepackage{tabularx}  
\usepackage{multirow}  
\usepackage{colortbl}
\usepackage{xcolor}
\usepackage{soul}
\definecolor{lightgreen}{RGB}{0,205,19}
\definecolor{PaleTurquoise}{RGB}{180,238,180}

\usepackage{enumitem} 
\usepackage{amssymb}

\usepackage[colorlinks=true, 
            linkcolor=customBlue, 
            citecolor=customBlue, 
            filecolor=magenta, 
            urlcolor=customBlue]{hyperref} 

\definecolor{customBlue}{RGB}{21, 0, 127} 
\definecolor{orange}{RGB}{236, 122, 16} 

\begin{document}

\title{Reassessing the Role of Chain-of-Thought in Sentiment Analysis: Insights and Limitations}

\titlerunning{Reassessing the Role of CoT in SA: Insights and Limitations}



\author{Kaiyuan Zheng\inst{1} \and
Qinghua Zhao\inst{2,3} \and
Lei Li\inst{3,4}}
\authorrunning{K. Zheng et al.}
%
\institute{Beijing Normal University, Zhuhai Campus, Zhuhai, China \\
\email{zhky@mail.bnu.edu.cn} \and
SAIBD, Hefei University, Hefei, China \\
SKLSDE, Beihang University, Beijing, China \\
\email{zhaoqh@buaa.edu.cn} \and
University of Washington, Seattle, USA \\
University of Copenhagen, Copenhagen, Denmark \\
\email{lilei@di.ku.dk}}

\maketitle      

\vspace{-12pt}

\begin{abstract}

The relationship between language and thought remains an unresolved philosophical issue.  Existing viewpoints can be broadly categorized into two schools: one asserting their independence, and another arguing that language constrains thought. In the context of large language models, this debate raises a crucial question: Does a language model's grasp of semantic meaning depend on thought processes? To explore this issue, we investigate whether reasoning techniques can facilitate semantic understanding. Specifically, we conceptualize thought as reasoning, employ chain-of-thought prompting as a reasoning technique, and examine its impact on sentiment analysis tasks. The experiments show that chain-of-thought has a minimal impact on sentiment analysis tasks. Both the standard and chain-of-thought prompts focus on aspect terms rather than sentiment in the generated content. Furthermore, counterfactual experiments reveal that the model's handling of sentiment tasks primarily depends on information from demonstrations. The experimental results support the first viewpoint.

\end{abstract}

\section{Introduction}
\vspace{-2pt}

In the realms of linguistics and cognitive science, the relationship between language and thought has long been a subject of profound inquiry and debate. Two contrasting viewpoints have emerged in this discourse. \cite{fedorenko2024language} argues for the independence of language and thought, positing that language serves merely as a vessel for thought, with each entity distinct and separate. In stark contrast, \cite{wittgenstein2023tractatus} proposes a more intricate relationship, suggesting that ``the limits of my language mean the limits of my world''. This perspective implies that the scope of our thoughts is fundamentally constrained by the language we possess to express them. The tension between these divergent views raises a critical question in the context of contemporary artificial intelligence: To what extent a language model's capacity is to grasp semantic meaning underlying thought processes?

\begin{figure}[!ht]
    \centering
    \includegraphics[width=0.75\linewidth]{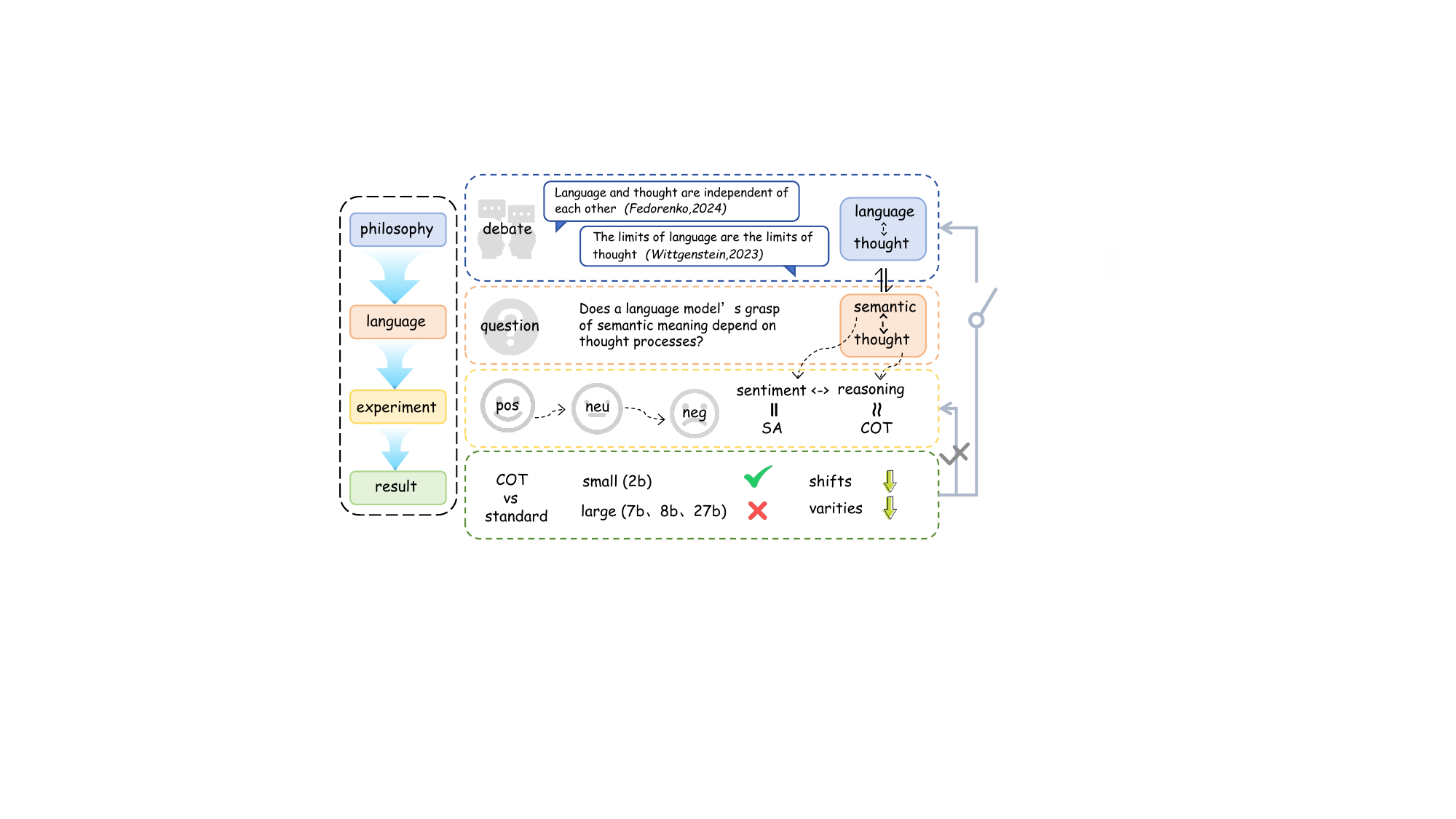}
    \caption{The framework of our work.}
    \vspace{-15pt}
    \label{fig:1}
\end{figure}

To answer the aforementioned question, we propose an experimental approach that examines tasks not explicitly reliant on reasoning abilities and evaluates whether providing additional reasoning information enhances model performance. The overall framework of our work is shown in Figure~\ref{fig:1}. We focus on aspect-based sentiment analysis (ABSA), a task that requires predicting the sentiment of specific aspect terms within reviews containing multiple aspect terms with varying sentiments. We integrate the chain-of-thought (CoT) prompting method to stimulate the model's reasoning capabilities in this context.

Our approach conceptualizes sentiment evolution as the reasoning path and the overall sentiment of the review as the reasoning outcome. Through this lens, we explore the relationship between language and thought. This indirect exploration is grounded in the following interconnections:

\vspace{-2pt}
\begin{itemize}
    \item Sentiment Analysis and Language: Sentiment analysis (SA) tasks fundamentally rely on the direct comprehension of language.
    \item Reasoning and Thought: The CoT method guides the model to generate answers through step-by-step reasoning, thereby stimulating the model's reasoning ability. We posit that this reasoning ability is a concrete manifestation of thought.
    \item Sentiment Analysis and Reasoning: Drawing from psychological ``Cognitive Load Theory'', we hypothesize that activating reasoning abilities helps achieve a deeper sentiment understanding.
\end{itemize}
\vspace{-2pt}

We selected two widely-used public ABSA datasets. Recognizing the limitations of these datasets in terms of emotional complexity, granularity, and dynamism, we also manually constructed a more nuanced emotion dataset. We designed three CoT formats, incorporating both natural language and symbolic language.
After analyzing the adaptation of CoT for SA tasks, we constructed a more complex dataset with diverse emotions and shifts to explore CoT's role in semantic understanding. We found that CoT has minimal impact on sentiment-oriented semantic tasks. To gain deeper insights, we further analyzed the attention changes between model inputs and outputs and explored whether the model's semantic understanding of sentiments stems from pre-training or demonstrations.

\section{Related work}


This paper involves analyzing SA tasks using CoT to determine whether the task leverages the model's reasoning ability, touching on reasoning (one kind of thought) and language. Therefore, the related work includes language and thought, chain-of-thought, and sentiment analysis.

\vspace{3pt}
\textbf{Language and Thought.} \cite{fedorenko2024language} found that language and thought are dissociated in the human brain. They discovered a schematic representation of the response profile of the language network (for example, as measured by fMRI). This network responds strongly to language comprehension and production but not to non-linguistic tasks that require thinking and reasoning. Therefore, they argue that language is a tool for communication, not for thinking, and that there is a clear distinction between the language system and various systems involved in thinking and reasoning. However, according to Wittgenstein, his famous idea, ``The limits of my language mean the limits of my world'', can be interpreted as thought being constrained by the structure and scope of language~\cite{wittgenstein2023tractatus}. If we cannot express something in language, we cannot fully grasp or conceptualize it in thought. This notion implies that language doesn't just communicate thoughts but also forms the boundaries of our cognitive processes.

\vspace{3pt}
\textbf{Chain-of-Thought.} CoT is an advanced form of in-context learning ~\cite{brown2020language}, designed to guide language models in generating coherent sequences of intermediate reasoning steps ~\cite{wei2022chain}. By providing step-by-step problem-solving processes in exemplars, CoT aims to lead models towards more accurate and justifiable answers to complex questions. This approach is particularly relevant to tasks requiring higher-level cognitive abilities, bridging the gap between language processing and thought-based reasoning.

\vspace{3pt}
\textbf{Interpretability in Sentiment Analysis.} Recent research has extensively explored the performance of large language models (LLMs) in SA tasks. \cite{ZHAO2023110792} highlighted the critical role of emotionally charged adjectives in determining overall sentiment, while \cite{wei2023largerlanguagemodelsincontext} investigated whether LLMs rely more on pre-trained knowledge or in-context exemplars when addressing SA tasks.
\cite{liu2023towards} utilized the SST-2 sentiment analysis benchmark to elucidate in-context learning mechanisms through the construction of contrastive examples. Notably, \cite{fei2023reasoning} pioneered the application of CoT to implicit SA, demonstrating the significant role of reasoning in this domain.
Further applications of CoT in SA include the work of ~\cite{rusnachenko2024nicolay}, who employed CoT to address emotion states and causes in conversations using six basic emotions. Similarly, \cite{lai2024rvisa} integrated CoT-style prompts into ABSA, consolidating reasoning steps within single exemplars.
Despite these advancements, intriguing findings by \cite{wang2018glue} and \cite{zhao2024wordorderworldknowledge} revealed that shuffling word order in SA tasks results in only marginal performance drops. This also raises another critical question: if word order has limited impact on SA, to what extent is reasoning ability necessary for these tasks?

\begin{table*}[!t]
    \centering
    \caption{Illustrative examples from the Laptop dataset. \textit{Italics} represent the aspect, and \sethlcolor{PaleTurquoise}\hl{green} highlight represents its sentiment.}
    \label{tab:sentiment_example_structure}
    \resizebox{\textwidth}{!}{%
    \begin{tabularx}{\textwidth}{l X} 
        \specialrule{1.2pt}{0pt}{0pt} 
        \textbf{Laptop} & \textbf{Examples} \\
        \specialrule{0.5pt}{0pt}{0pt} 
        \textbf{Explicit} & \textbf{Sentence:} Overall I feel this netbook was \sethlcolor{PaleTurquoise}\hl{poor} \textit{quality}, had \sethlcolor{PaleTurquoise}\hl{poor} \textit{performance}, although it did have \sethlcolor{PaleTurquoise}\hl{great} \textit{battery life} when it did work. \\
        & \textbf{Overall Sentiment:} Negative \\
        \textbf{Implicit} & \textbf{Sentence:} Also, in using the \textit{built-in camera}, my \textit{voice recording} for my vlog sounds like interplanetary transmissions in the ``Star Wars'' saga. \\
        & \textbf{Overall Sentiment:} Negative \\
        \specialrule{1.2pt}{0pt}{0pt} 
    \end{tabularx}
    }
    \vspace{-10pt}
\end{table*}

\section{Experimental Setup}

This section outlines the LLMs, datasets, and prompts employed in our experimental framework.

\subsection{Models}
Our experiments utilize a range of models varying in size and architecture, including Gemma-2 (2B, 9B, 27B) ~\cite{team2024gemma} and LLaMA-3 8B ~\cite{dubey2024llama}. These models were deployed on two A800 GPUs, each equipped with 80GB of memory, and operated in float32 precision to ensure optimal performance and accuracy.

\subsection{Datasets}
For our analysis, we selected two widely recognized ABSA datasets from SemEval-2014 ~\cite{pontiki2016semeval}: the Laptop and Restaurant datasets. 
Refer to Table~\ref{tab:sentiment_example_structure} for examples.

To align more closely with our research objectives, we applied a set of criteria to select test samples. 
\begin{itemize}
    \item Text length: We prioritized longer text samples to ensure comprehensive semantic expression and sufficient scope for sentiment shifts.
    \item Sentiment dynamics: Selected samples exhibit sentiment changes to assess the model's capacity for understanding aspect sentiment.
    \item Complexity: Each sample contains a minimum of two aspects and two sentiment changes to ensure sufficient complexity.
    \item Sentiment split: Following ~\cite{li2021learning}, we categorized the samples into explicit and implicit splits. Explicit data contains direct expressions of sentiment or emotion, where the sentiment is clearly articulated (e.g., ``Just ten minutes away from you makes me want to cry''). In contrast, implicit data captures more nuanced cues where sentiment is indirectly conveyed through context or subtler language (e.g., ``I miss you''). 
\end{itemize}

\subsection{CoT-style prompts}\label{sec:prompt}
For standard prompts, we directly construct it by concatenating the input question and answers. 
For CoT-style prompts,  we require them to describe each aspect sentiment one by one. 
Different CoT-style prompts can lead to significant performance differences. Even when prompts are semantically similar, LLMs may generate vastly different responses~\cite{lu2021fantastically,perez2021true}. Therefore, to avoid the experimental conclusions being biased by a specific prompt,we tested three different versions of the CoT strategy, covering various levels from natural language expression to symbolic representation.

Specifically, the first version we used is a purely natural language-based CoT, which relies entirely on natural language to express the reasoning process for sentiment polarity. This version aims to simulate the sentiment reasoning process used in everyday human language, emphasizing the naturalness and coherence of language, called CoT-{\color{lightgreen} v1}. The third version is a symbol-based CoT, where symbols and logical expressions are used to describe sentiment polarity shifts, reducing the reliance on natural language and placing a stronger emphasis on the logical aspects, named CoT-{\color{lightgreen} v3}. Additionally, we employed a hybrid CoT, which strikes a balance between the two approaches, combining natural and symbolic language to balance the naturalness of language expression with the logical rigor of reasoning, named CoT-{\color{lightgreen} v2}. Examples of the different versions are shown in Table \ref{tab:cot-examples}.

\section{Experiments}

\vspace{-8pt}

This section delineates our experimental framework designed to address four pivotal research questions:

\begin{itemize}
    \item \textbf{RQ-1:} What is the adaptation of CoT on SA?
    \item \textbf{RQ-2:} Is it conflict or consistency with more complex emotions?
    \item \textbf{RQ-3:} How does CoT affect the correlation between input questions and output tokens?
    \item \textbf{RQ-4:} Does the model rely on knowledge acquired during the pre-training  or in the CoT exemplars?
\end{itemize}

\subsection{RQ-1: Adaptation of CoT in SA}
\begin{figure*}[!t]
    \centering
    \includegraphics[width=0.16\linewidth]{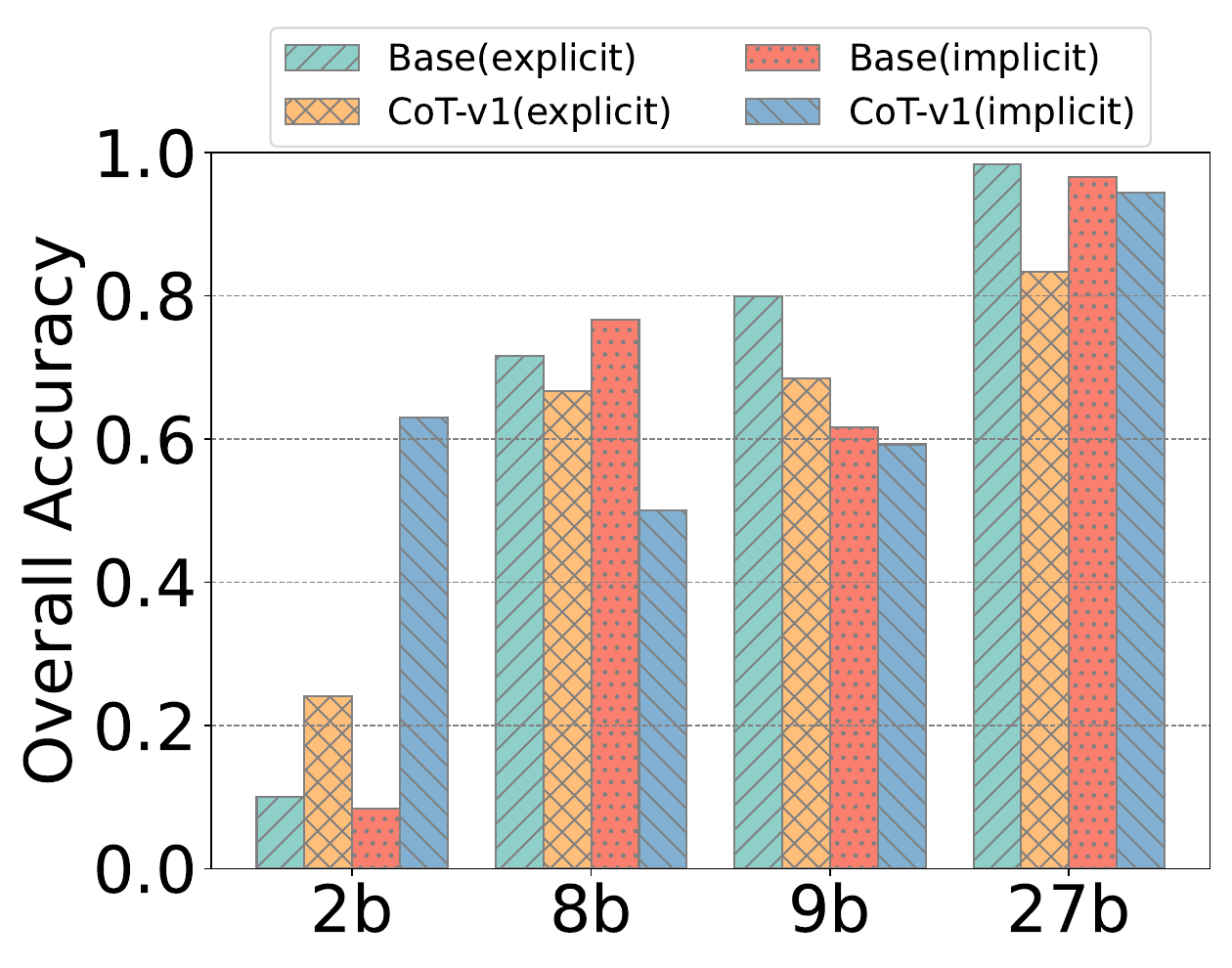} 
    \includegraphics[width=0.16\linewidth]{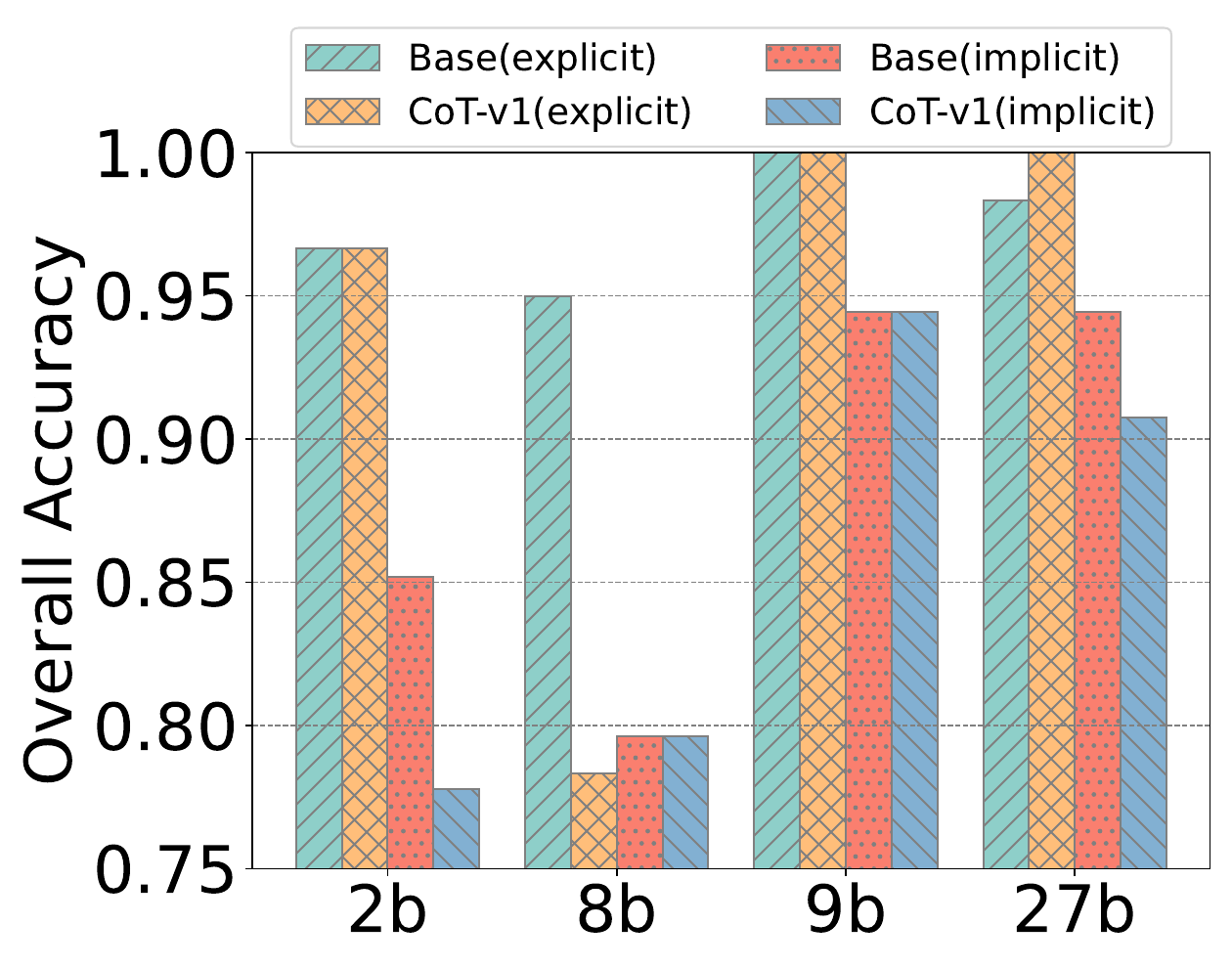} 
    \includegraphics[width=0.16\linewidth]{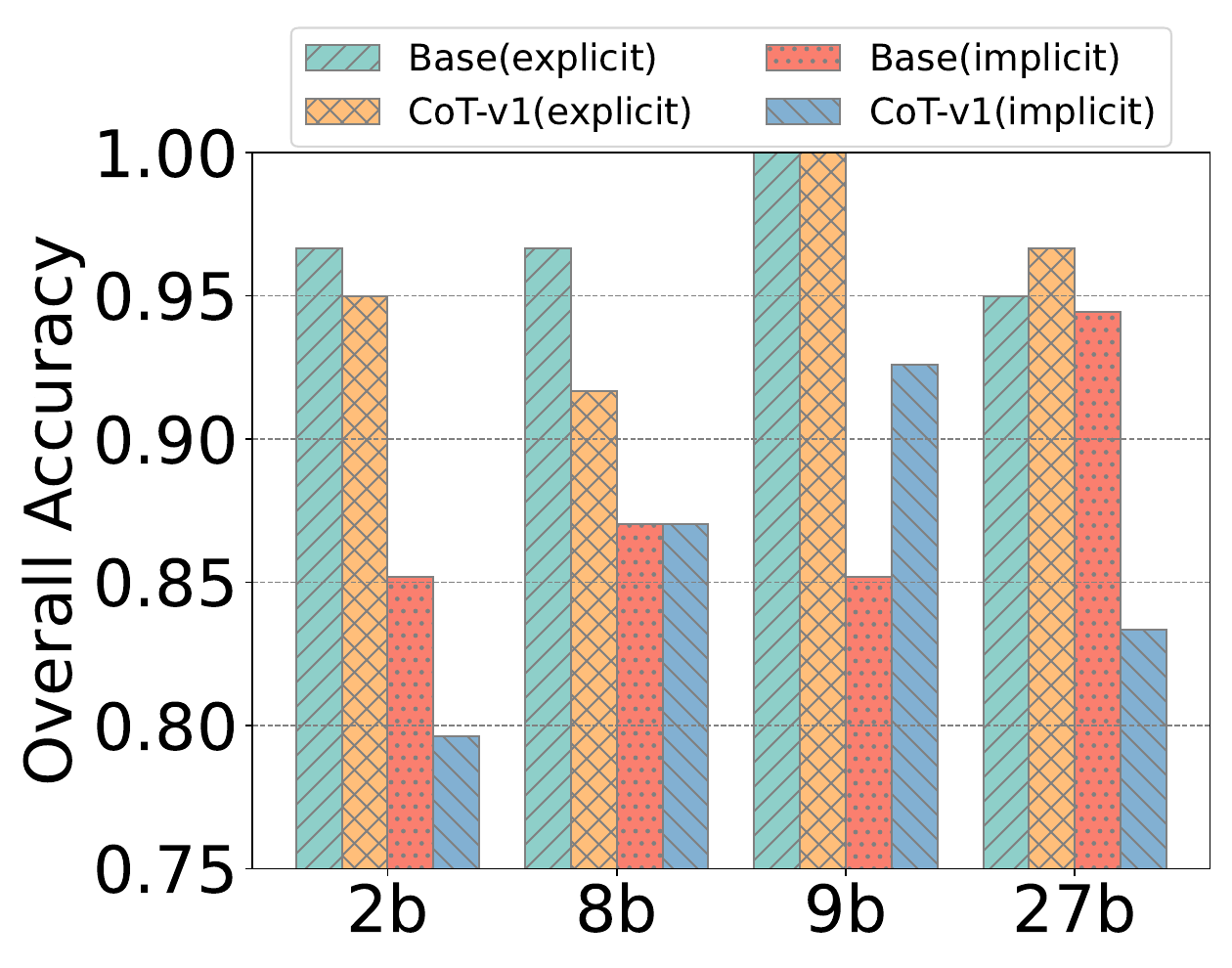} 
    \includegraphics[width=0.16\linewidth]{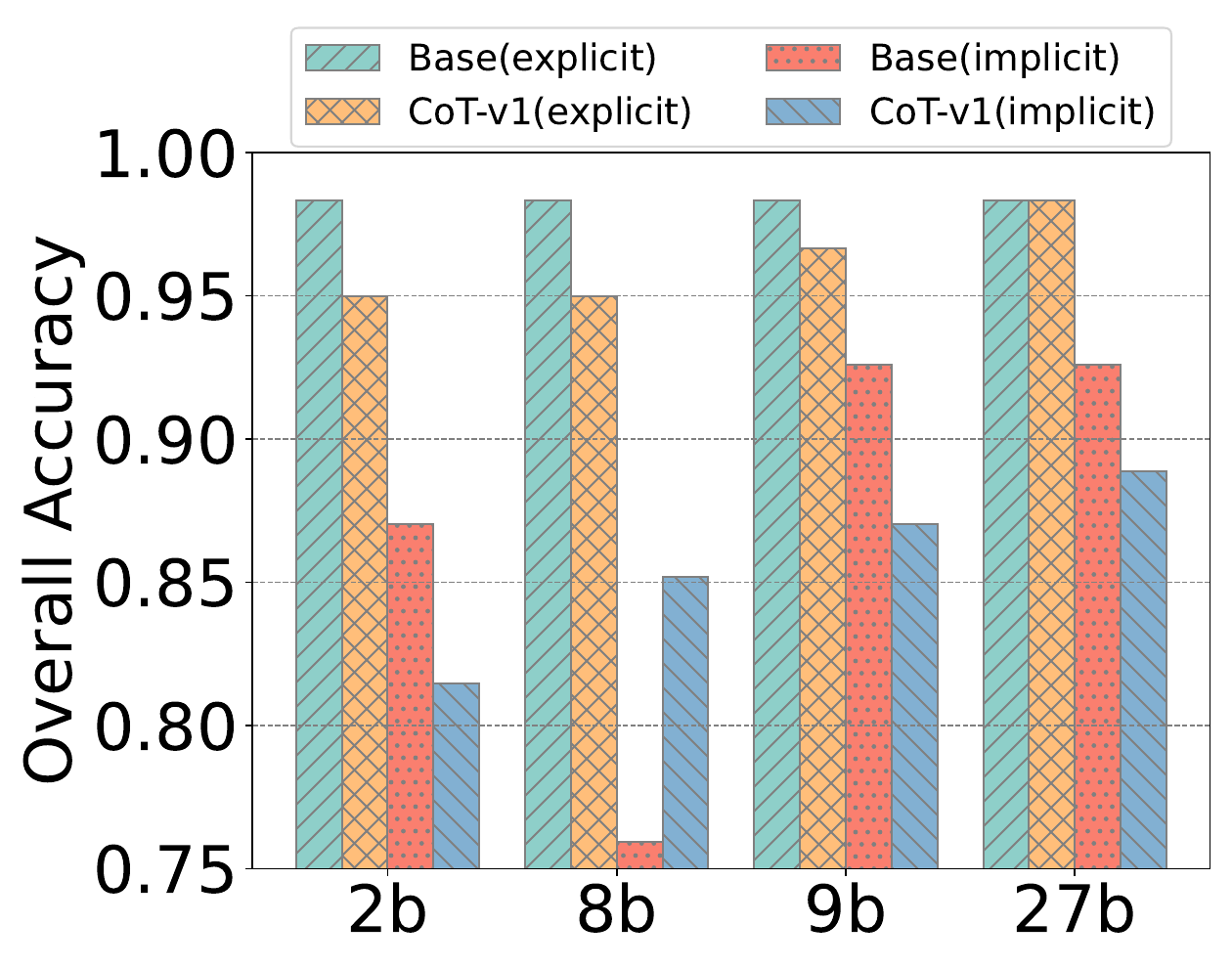} 
    \includegraphics[width=0.16\linewidth]{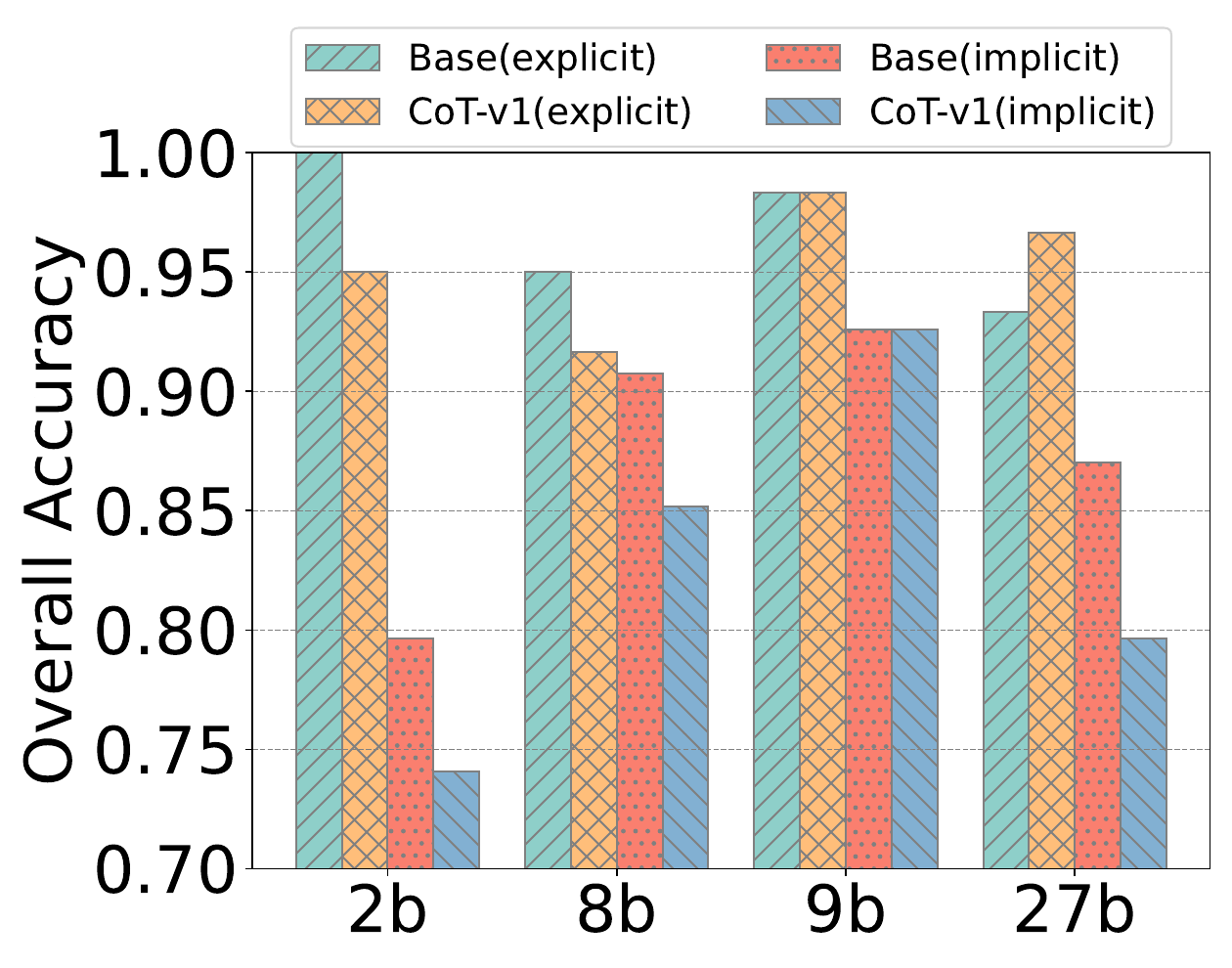} 
    \includegraphics[width=0.16\linewidth]{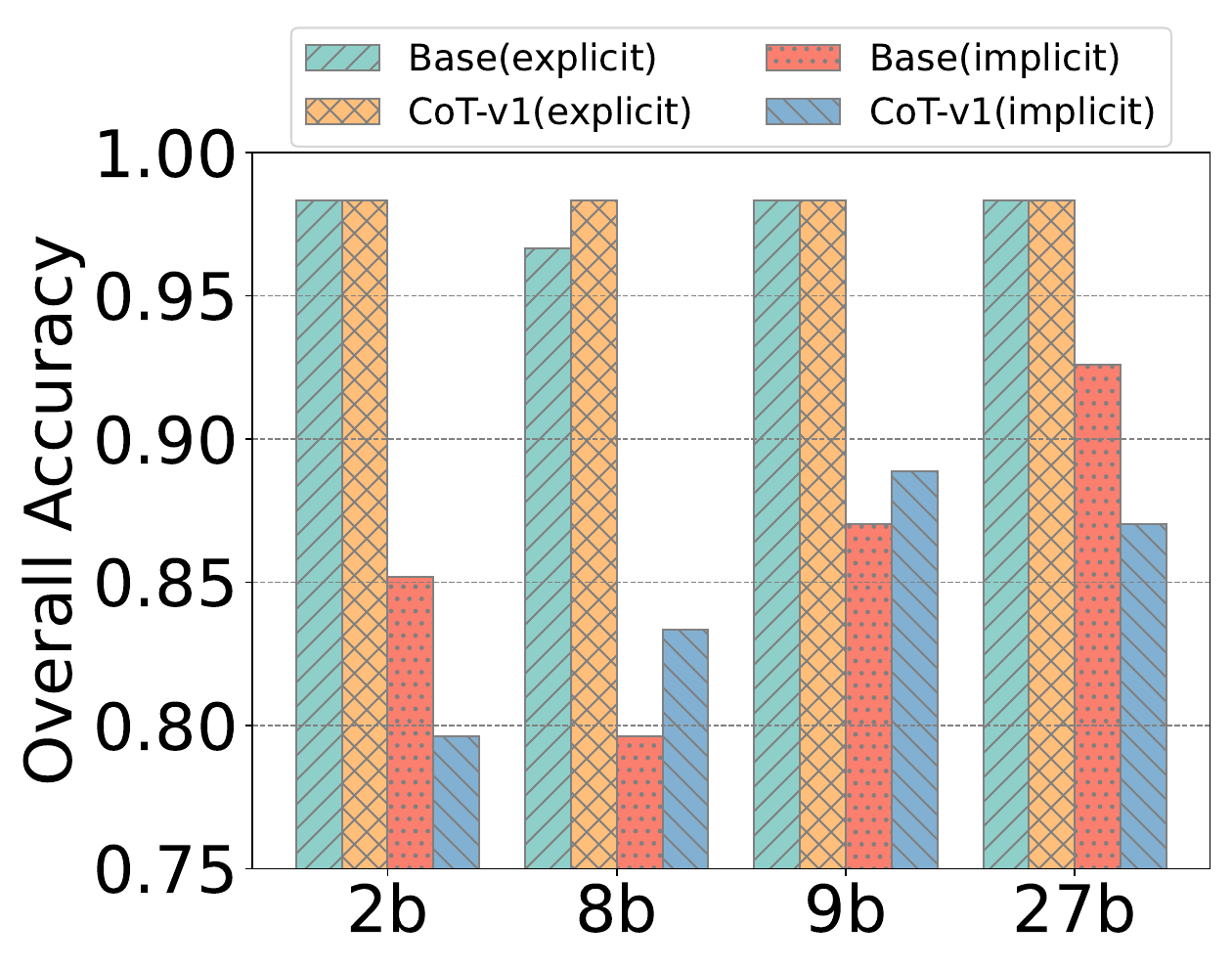} 

    \includegraphics[width=0.16\linewidth]{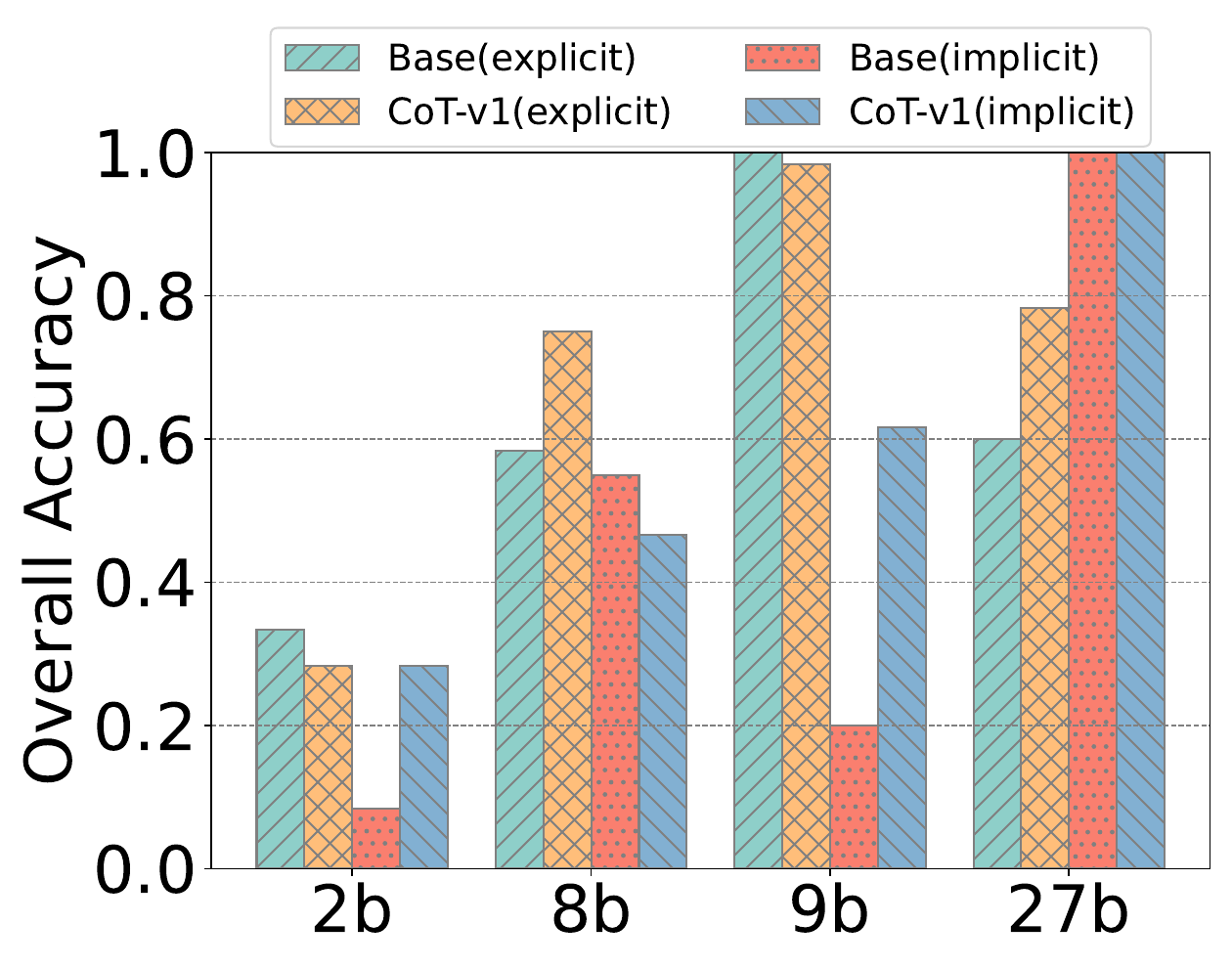} 
    \includegraphics[width=0.16\linewidth]{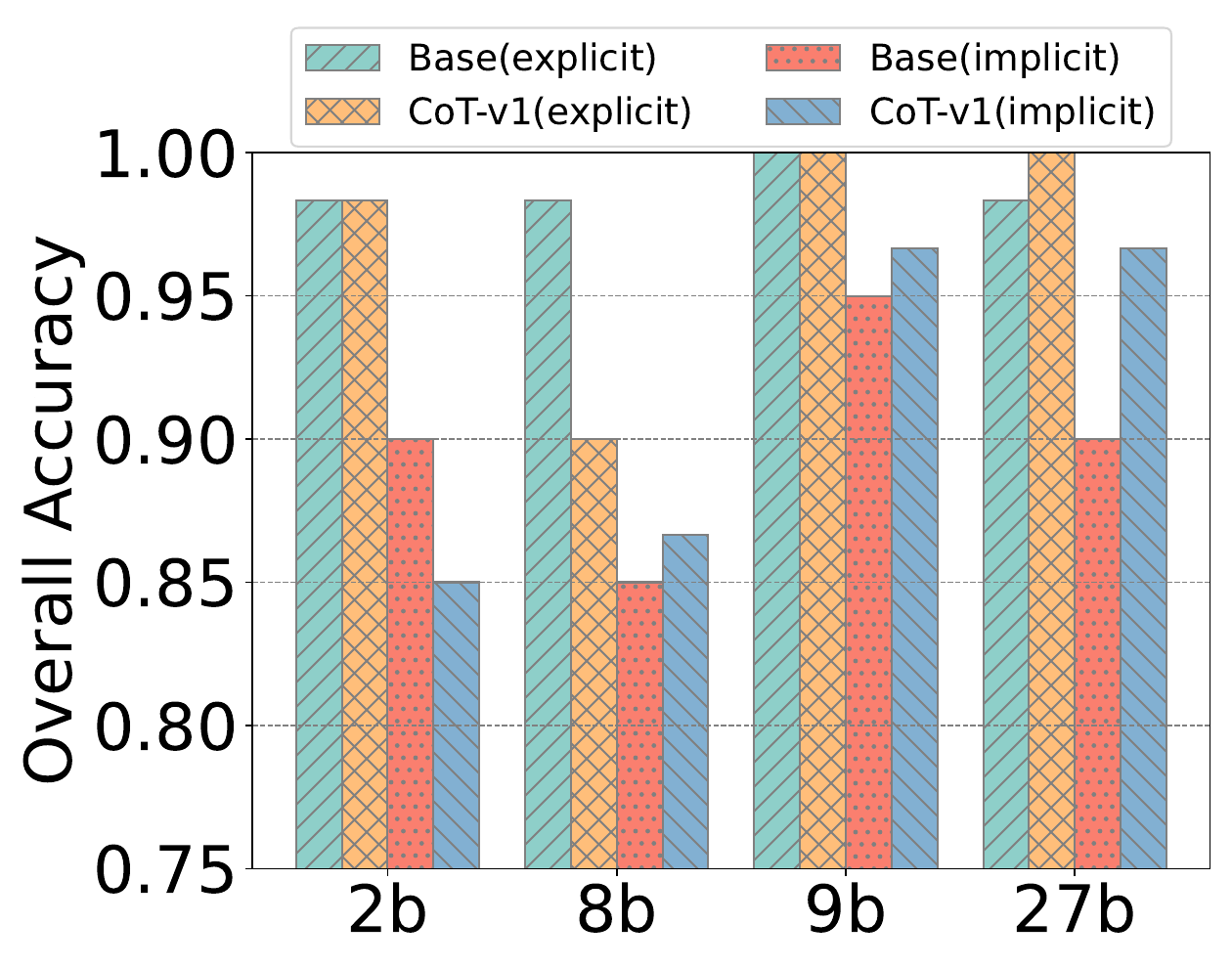} 
    \includegraphics[width=0.16\linewidth]{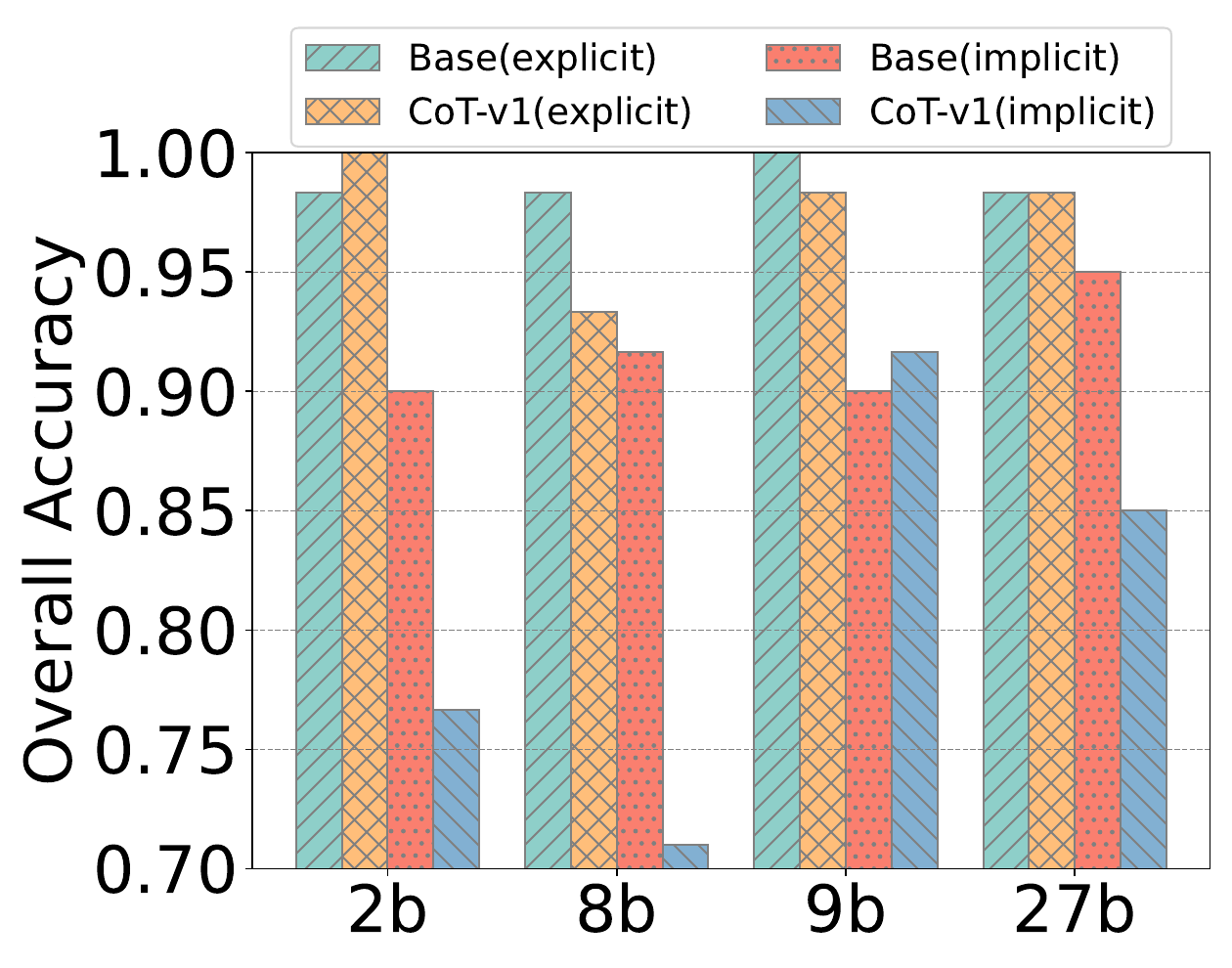} 
    \includegraphics[width=0.16\linewidth]{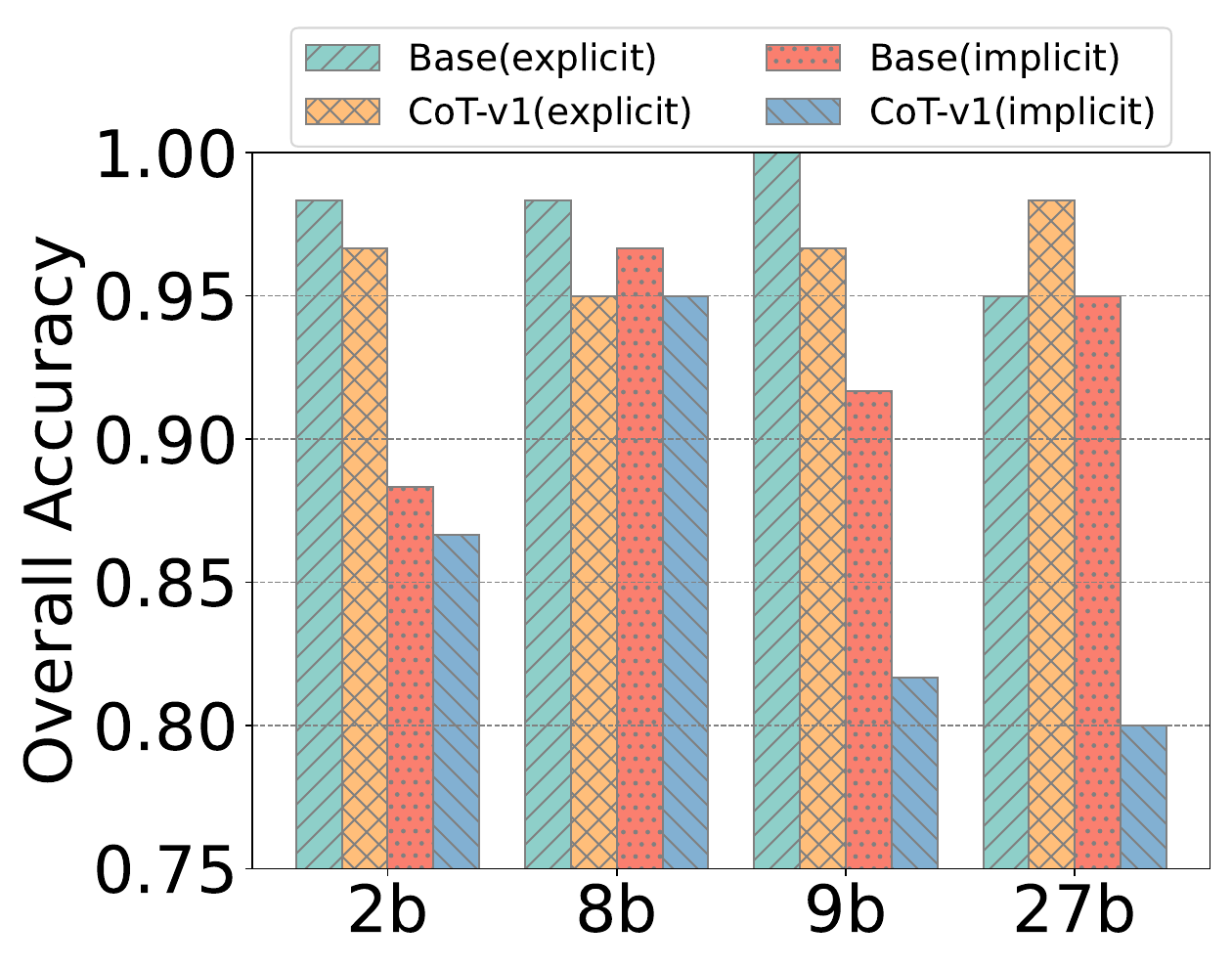} 
    \includegraphics[width=0.16\linewidth]{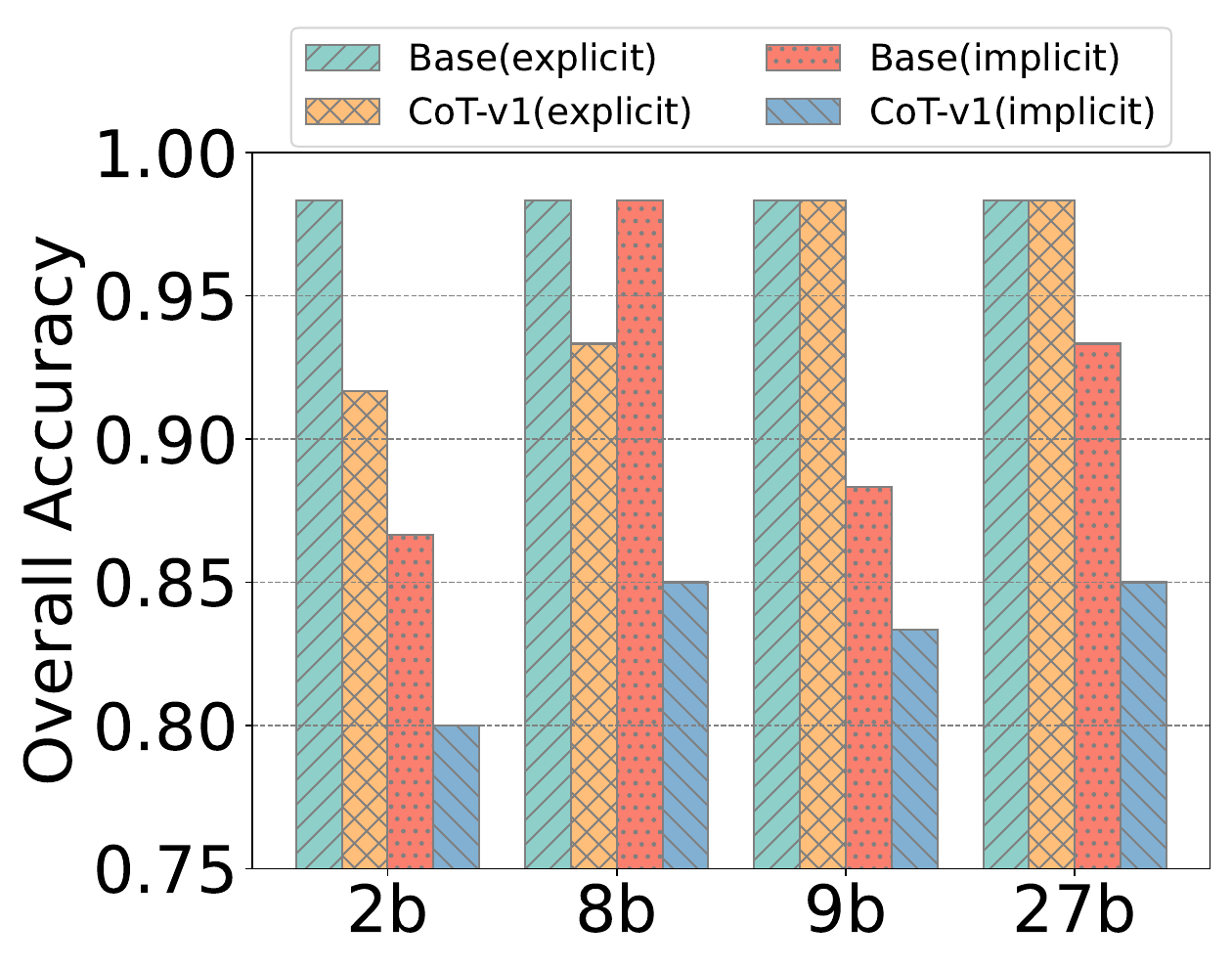} 
    \includegraphics[width=0.16\linewidth]{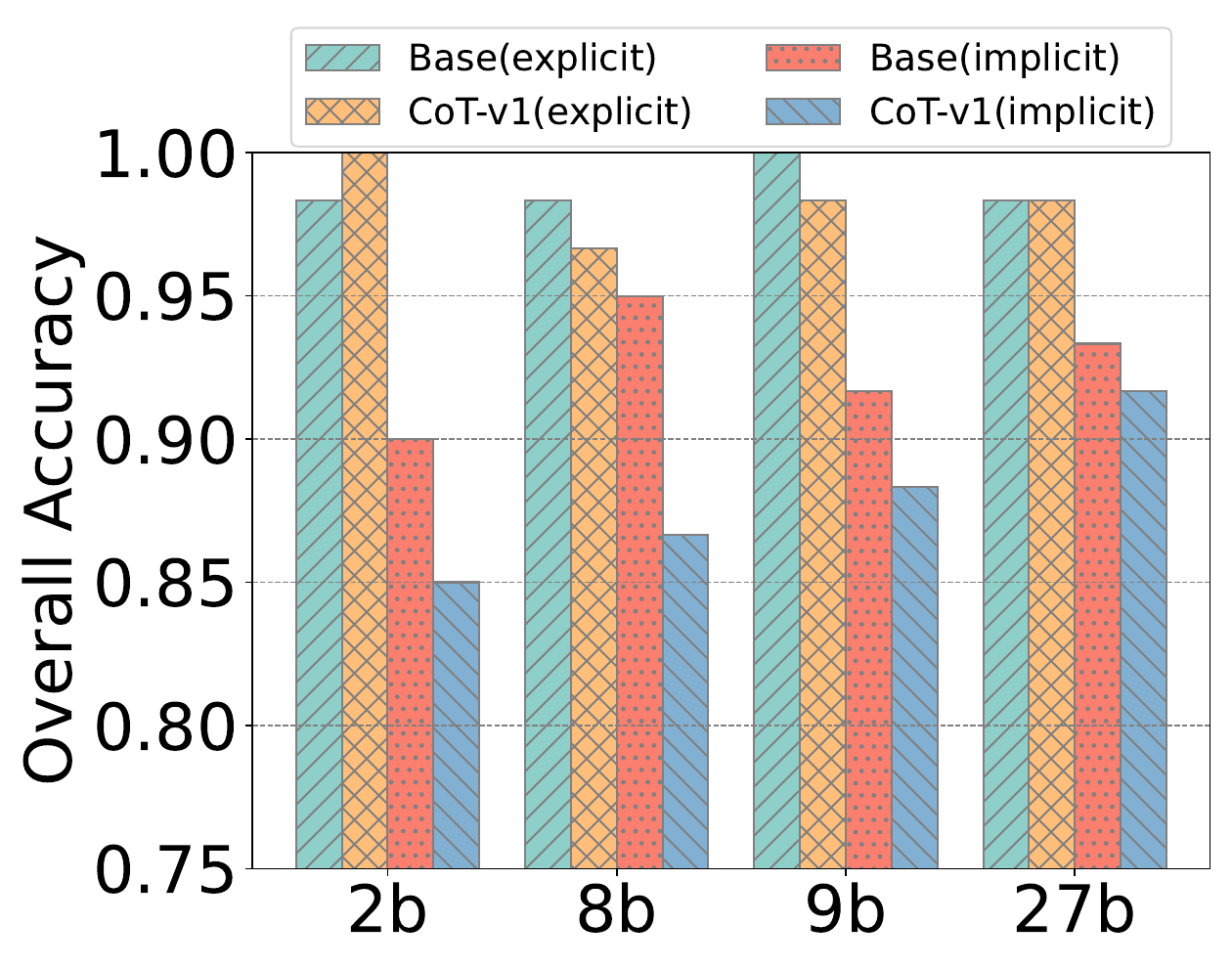} 
    \caption{Overall accuracy across datasets and shots. The top row presents results for the Laptop, while the bottom row shows the Restaurant. Each row displays accuracy values for 1-, 4-, 8-, 12-, 15-, and 18-shot settings, respectively.}
    \label{fig:overall_acc}
    \vspace{-10pt}
\end{figure*}

\vspace{-3pt}
To investigate whether reasoning techniques can enhance semantic understanding, we examined the impact of CoT on SA tasks. As shown in Figure~\ref{fig:overall_acc}, we reported results of CoT-v1 across four models, six shot settings, and two datasets with explicit and implicit splits.

Our findings reveal that: CoT yields improvements for the smallest model (Gemma2-2b) and in 1-shot scenarios. For instance, on the implicit split of Laptop dataset with Gemma2-2b and 1-shot, accuracy increased from {\color{orange} 0.24} (standard prompt) to {\color{orange} 0.62} (CoT-{\color{lightgreen} v1}). However, for larger models, CoT's impact on SA is minimal. This limited improvement may be attributed to the relative simplicity of SA tasks for current LLMs, which already achieve high accuracy (>{\color{orange} 0.95}) with standard prompts. Besides, as the number of demonstrations increases, CoT's effectiveness diminishes. For example, the improvement for Gemma2-27b on the explicit split of Restaurant dataset drops from {\color{orange} 0.18} (1-shot) to {\color{orange} 0.0} (18-shot).

It is important to note that our experimental design required an additional step due to the absence of ground-truth overall sentiment in the Laptop and Restaurant datasets. To address this, we implemented a post-hoc analysis using a weighted majority voting method to establish proxy ground-truths.

\begin{figure*}[!t]
    \centering
    \includegraphics[width=0.24\linewidth]{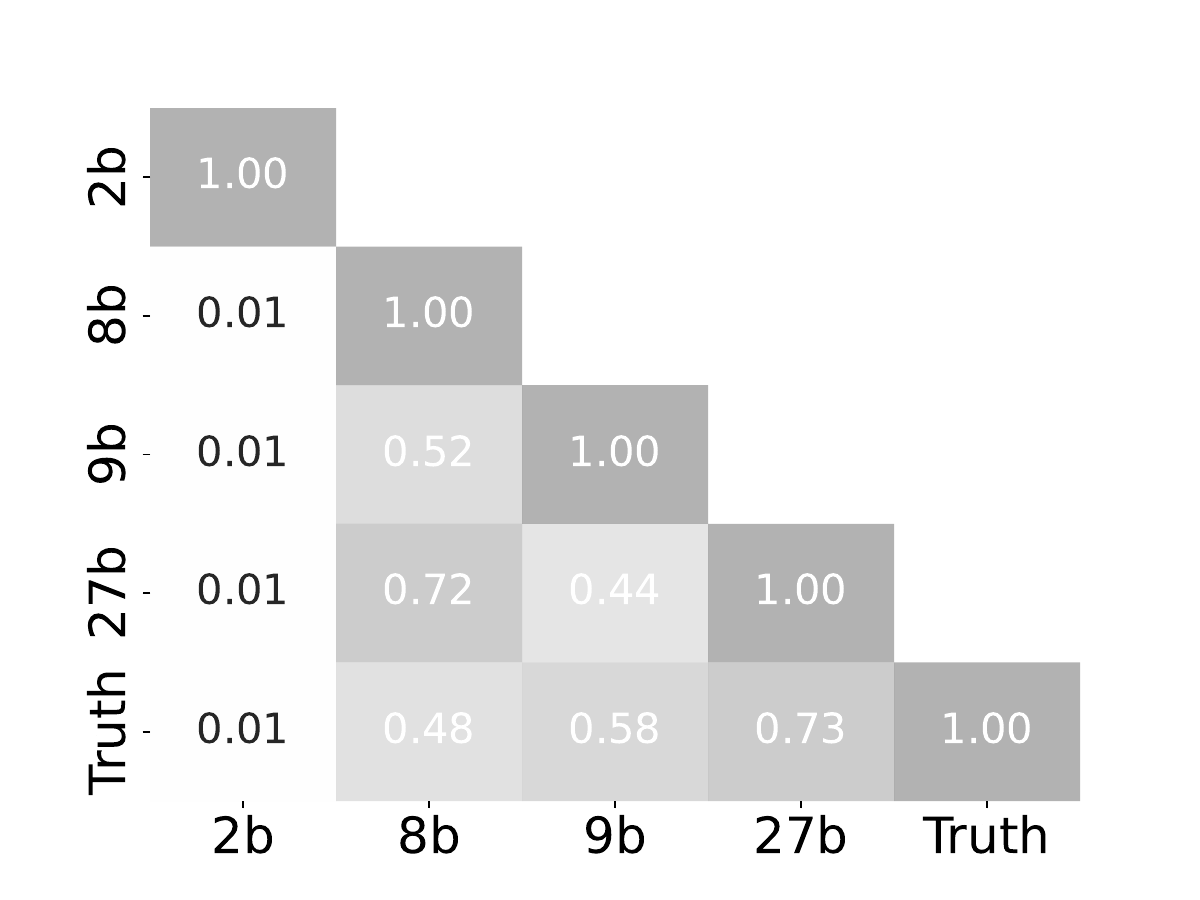}
    \includegraphics[width=0.24\linewidth]{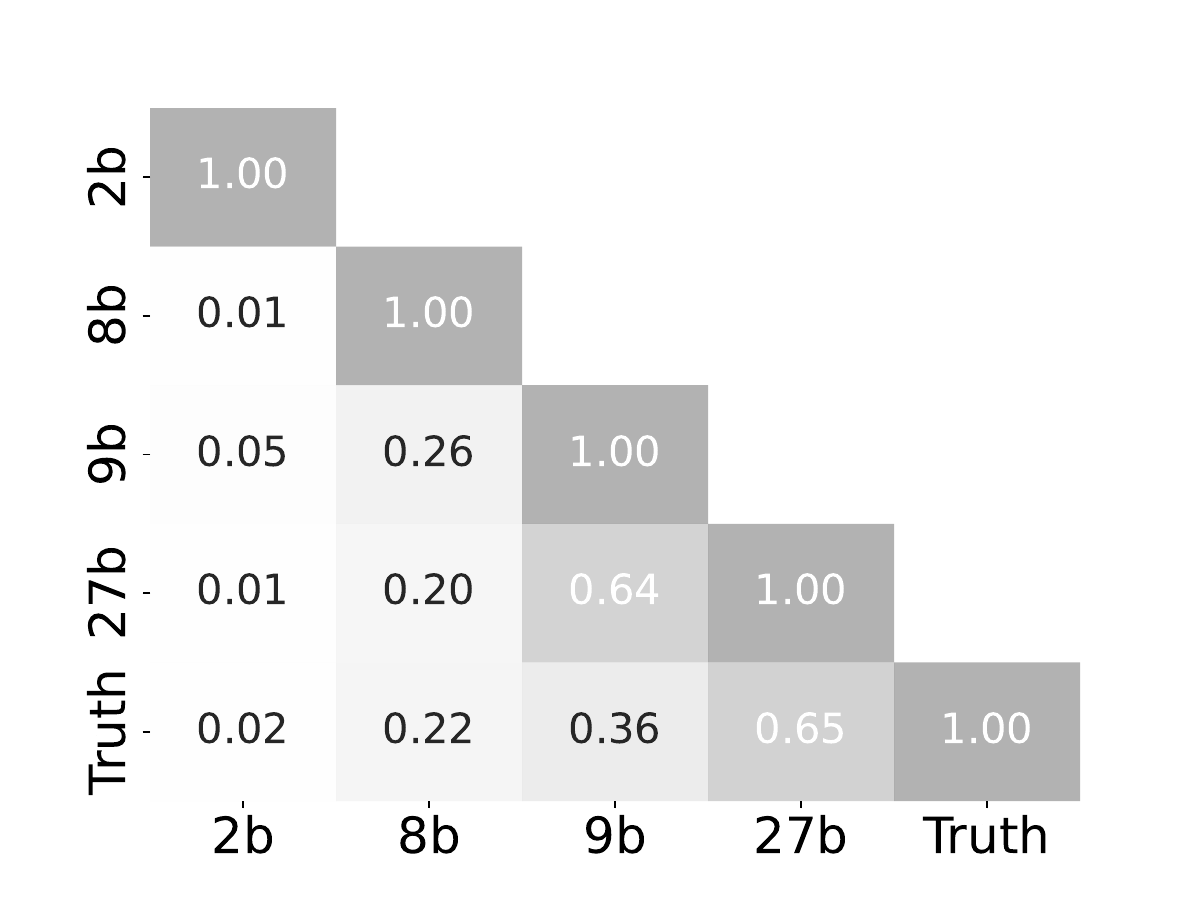}
    \includegraphics[width=0.24\linewidth]{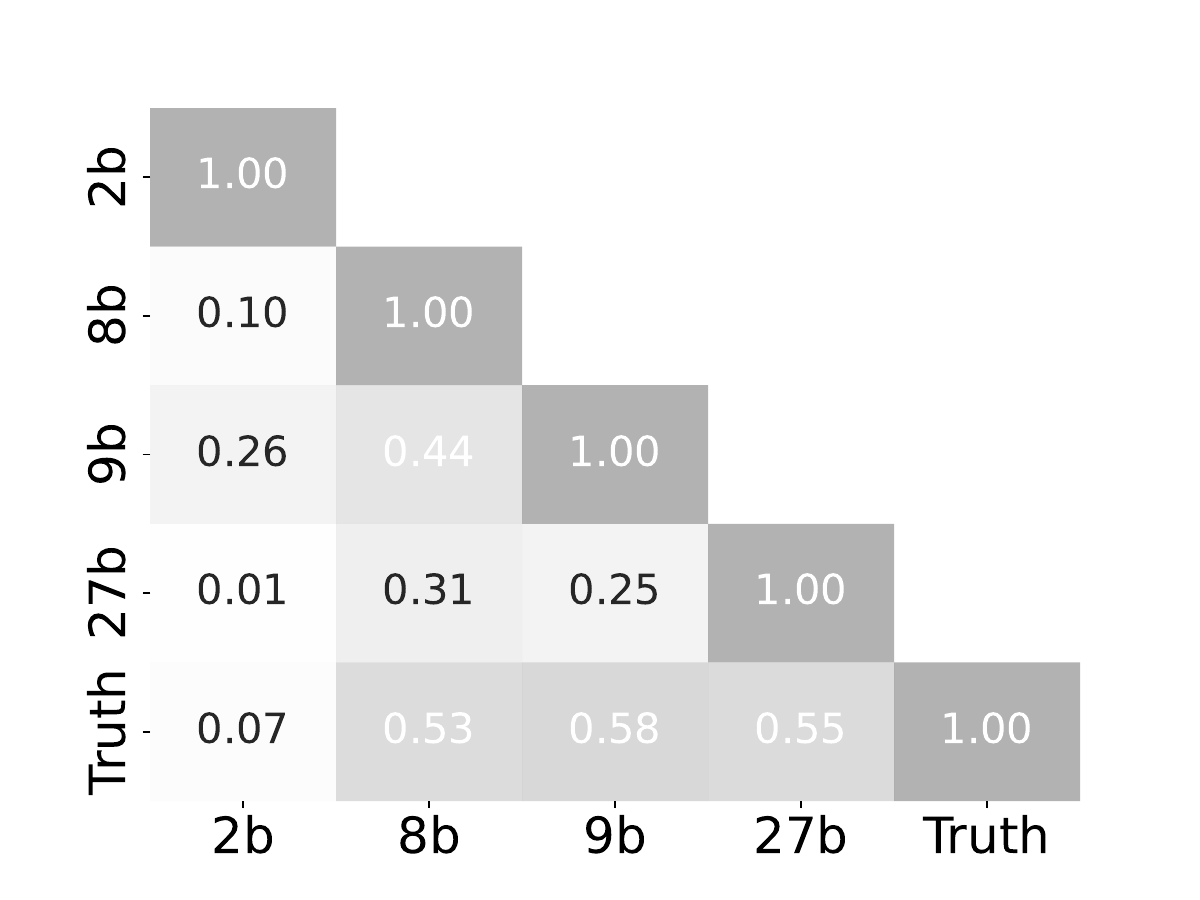}
    \includegraphics[width=0.24\linewidth]{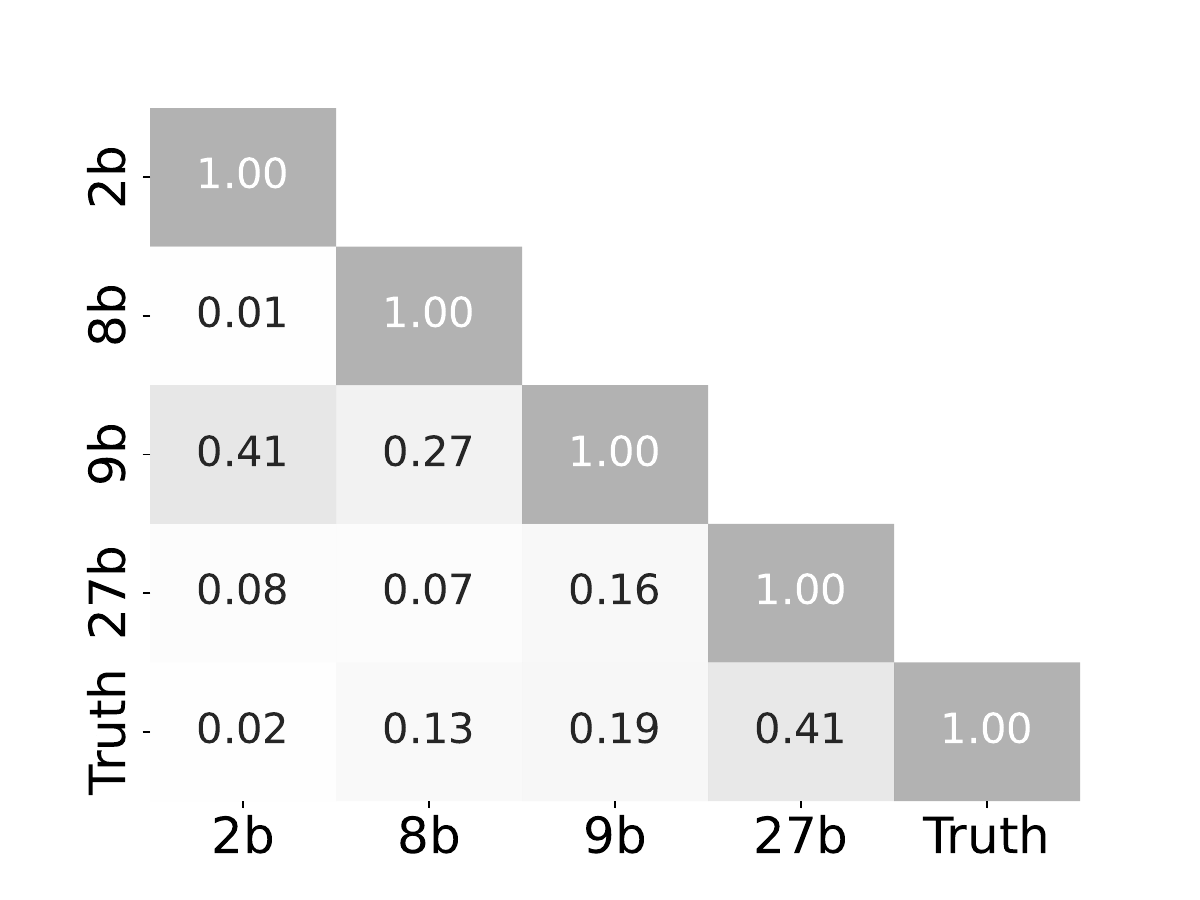}
    \caption{Agreement (Cohen’s Kappa value, treated as weights for majority voting) between model predictions and ground truth across datasets and sentiment types (1-shot setting). From left to right: Laptop (explicit), Laptop (implicit), Restaurant (explicit), and Restaurant (implicit).}
    \label{fig:consistency}
    \vspace{-10pt}
\end{figure*}

Figure \ref{fig:consistency} shows the agreements. It can be observed that larger models (such as Gemma2-27b) show higher agreement with true labels, whereas smaller models (such as Gemma2-2b) exhibit lower agreement, e.g, {\color{orange} 0.73} vs. {\color{orange} 0.01}. Furthermore, across different model sizes, implicit sentiment generally shows lower agreement with ground-truth compared to explicit sentiment, based on the agreement results from the above experiments, a weight is assigned to each model's voting. 

\subsection{RQ-2: Conflict or consistency?}
To further test the impact of CoT on SA tasks, we constructed a emotional analysis dataset with higher emotional complexity, yielding results consistent with the Laptop and Restaurant datasets. Additionally, we explored the effect of the number of emotion categories and the count of emotional shifts on model performance.

\textbf{Construction of  multi-emotion shift dataset.} To facilitate this investigation, we manually constructed a novel multi-emotion shift dataset (MES) featuring fine-grained emotional expressions. This dataset is characterized by texts containing multiple emotion types and frequent emotional shifts within single narratives.
Given the complexity and often overlapping nature of human emotions, we focused on six distinct major emotional categories: fear, happiness, anxiety, jealousy, loneliness, and shame. These emotions were contextualized within various scenarios including work environments, public transportation, entertainment activities, social interactions, and dining experiences.
We created 100 emotion-shift texts, each incorporating at least two emotional transitions. These narratives were crafted to reflect real-life situations, maintaining a balance between scenario continuity and cross-scenario coherence. This approach ensures both the representativeness of the dataset and its suitability for exploring the impact of emotional complexity on model reasoning capabilities.

\begin{figure}[!ht]
    \centering
    \vspace{-5pt}
    \includegraphics[width=0.32\linewidth]{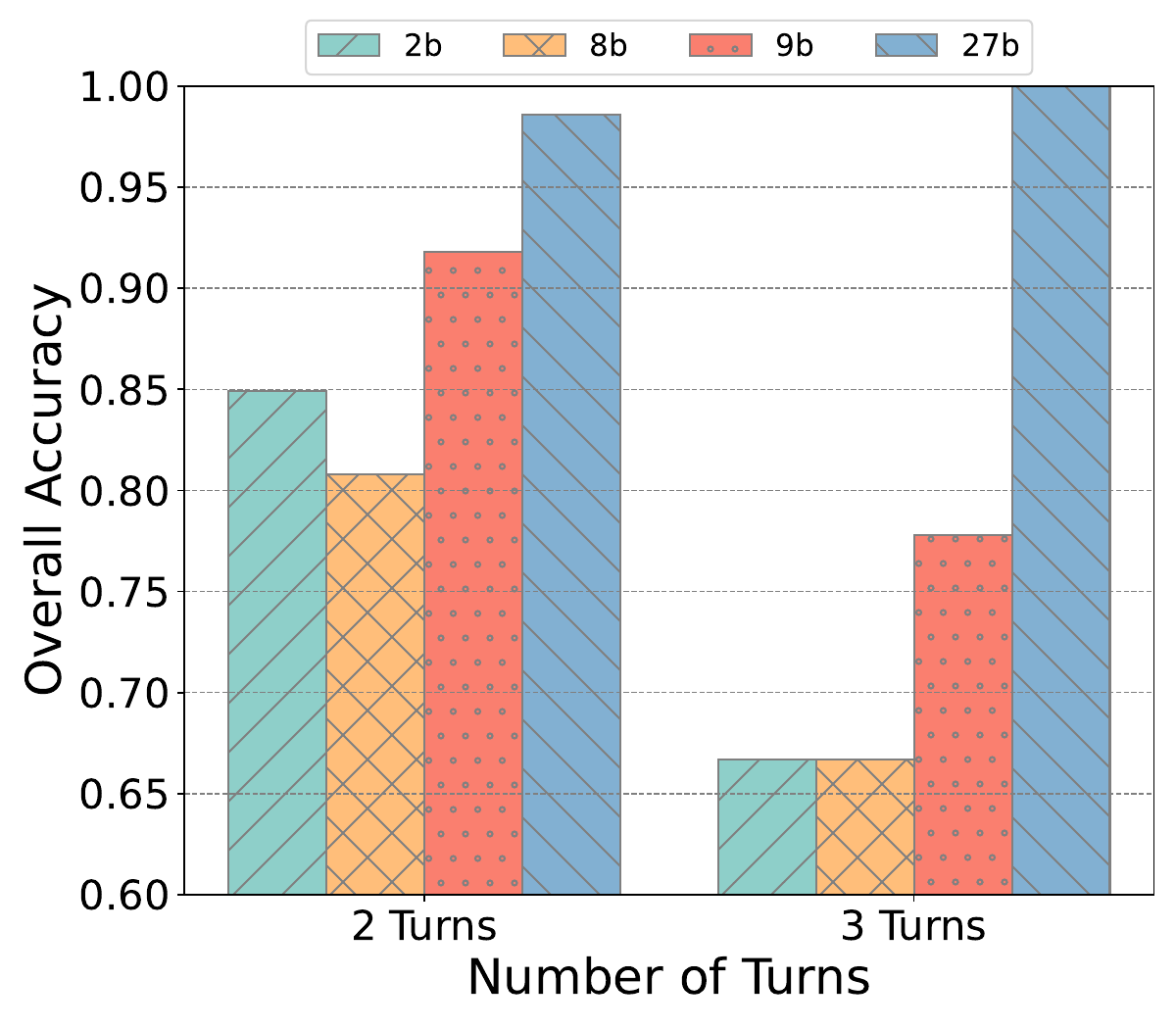}
    \includegraphics[width=0.32\linewidth]{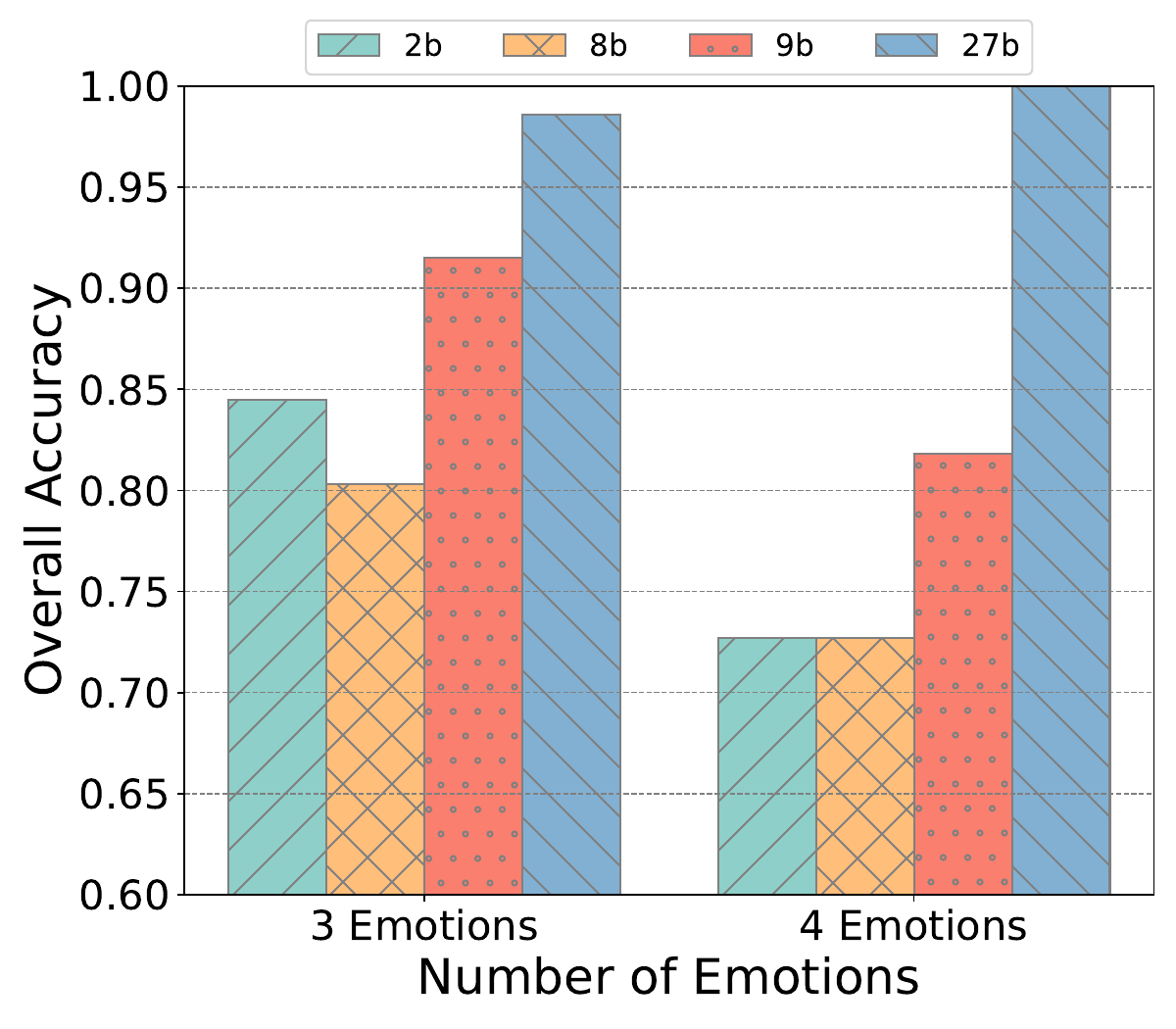}
    \caption{Model accuracy as a function of emotion shifts and categories. Results shown for 4-shot CoT-{\color{lightgreen} v1} on the MES dataset.}
    \vspace{-20pt}
    \label{fig:number}
\end{figure}

\textbf{Conflict increases difficulty. } Figure \ref{fig:number} illustrates the relationship between overall accuracy and emotional complexity, considering both the number of emotion categories and the frequency of emotional shifts within texts. Our findings reveal a consistent trend across most models:
\begin{itemize}
    \item  Models demonstrate lower accuracy when texts contain more frequent emotional shifts. For instance, Gemma2-9b's accuracy decreases from {\color{orange} 0.92} to {\color{orange} 0.78} as the number of emotional shifts increases from 2 to 3.
    \item  Similarly, accuracy declines with an increase in the number of emotion categories present in a text. Gemma2-9b shows a drop in accuracy from {\color{orange} 0.81} to {\color{orange} 0.73} when the number of emotion categories increases from 3 to 4.
\end{itemize}

These results suggest both the frequency of emotional shifts and the diversity of emotion categories contribute to the complexity of SA tasks. Texts with fewer emotional shifts and a more limited range of emotion categories (i.e., exhibiting greater emotional consistency) appear to be more manageable for the models.

\subsection{RQ-3: Correlation between Input and Output Tokens}
This section further explores how input interacts with output tokens when using CoT. We approach this question by analyzing the similarities between input questions and generated answers.



\begin{figure*}[!ht]
    \vspace{-10pt}
    \centering
    \begin{tabular}{ccc}
    \includegraphics[width=0.25\textwidth]{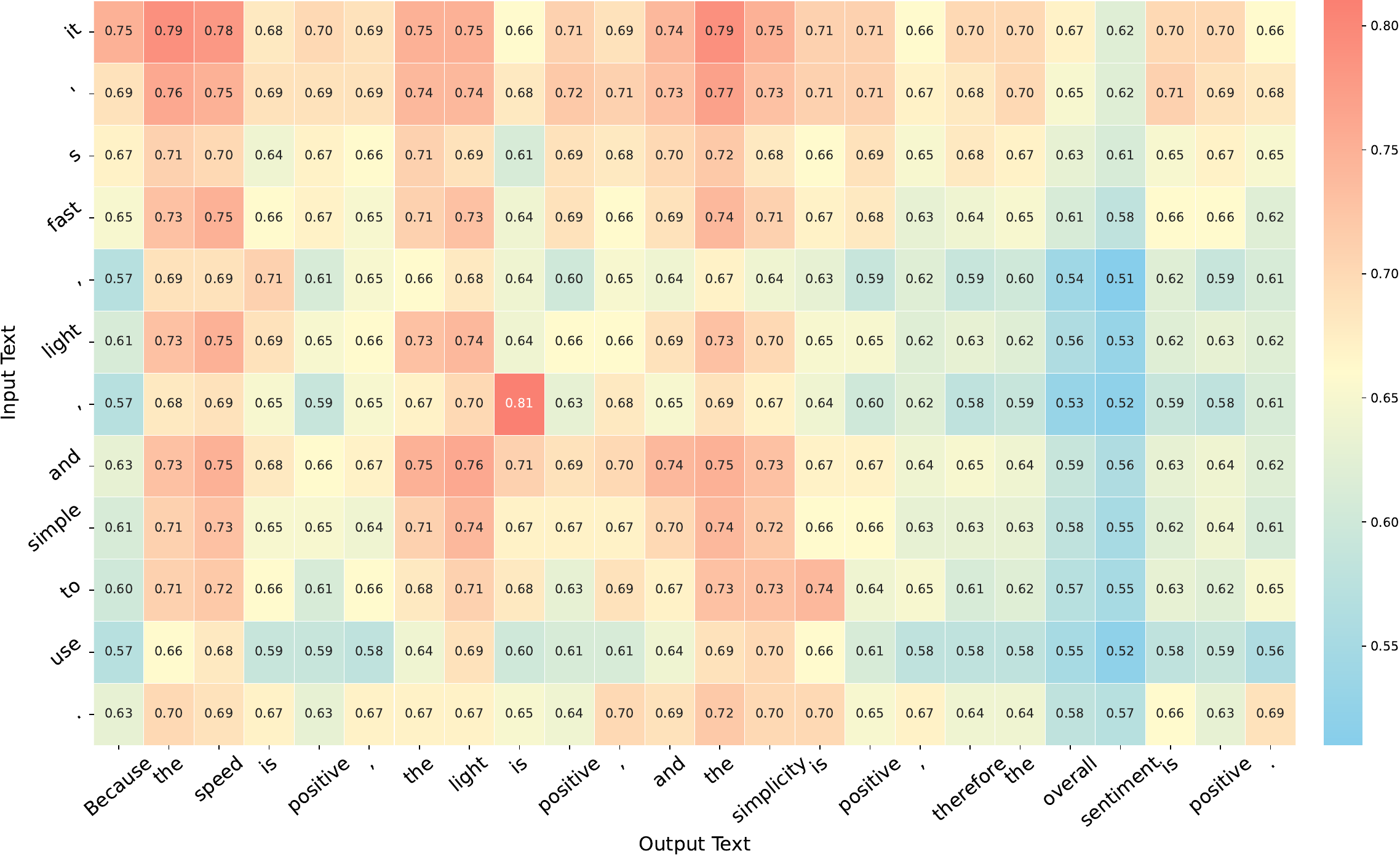}
    \includegraphics[width=0.25\textwidth]{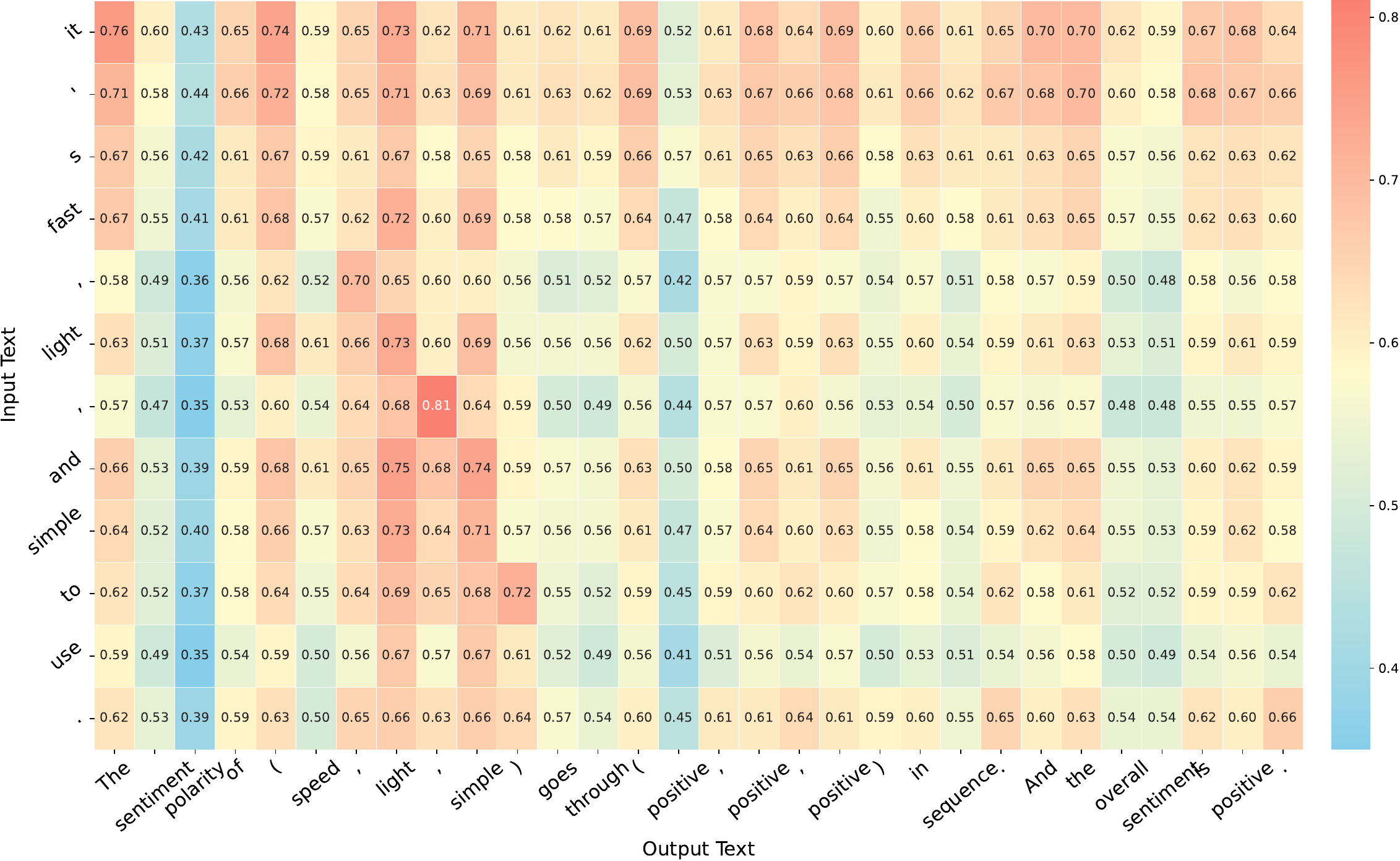}
    \includegraphics[width=0.25\textwidth]{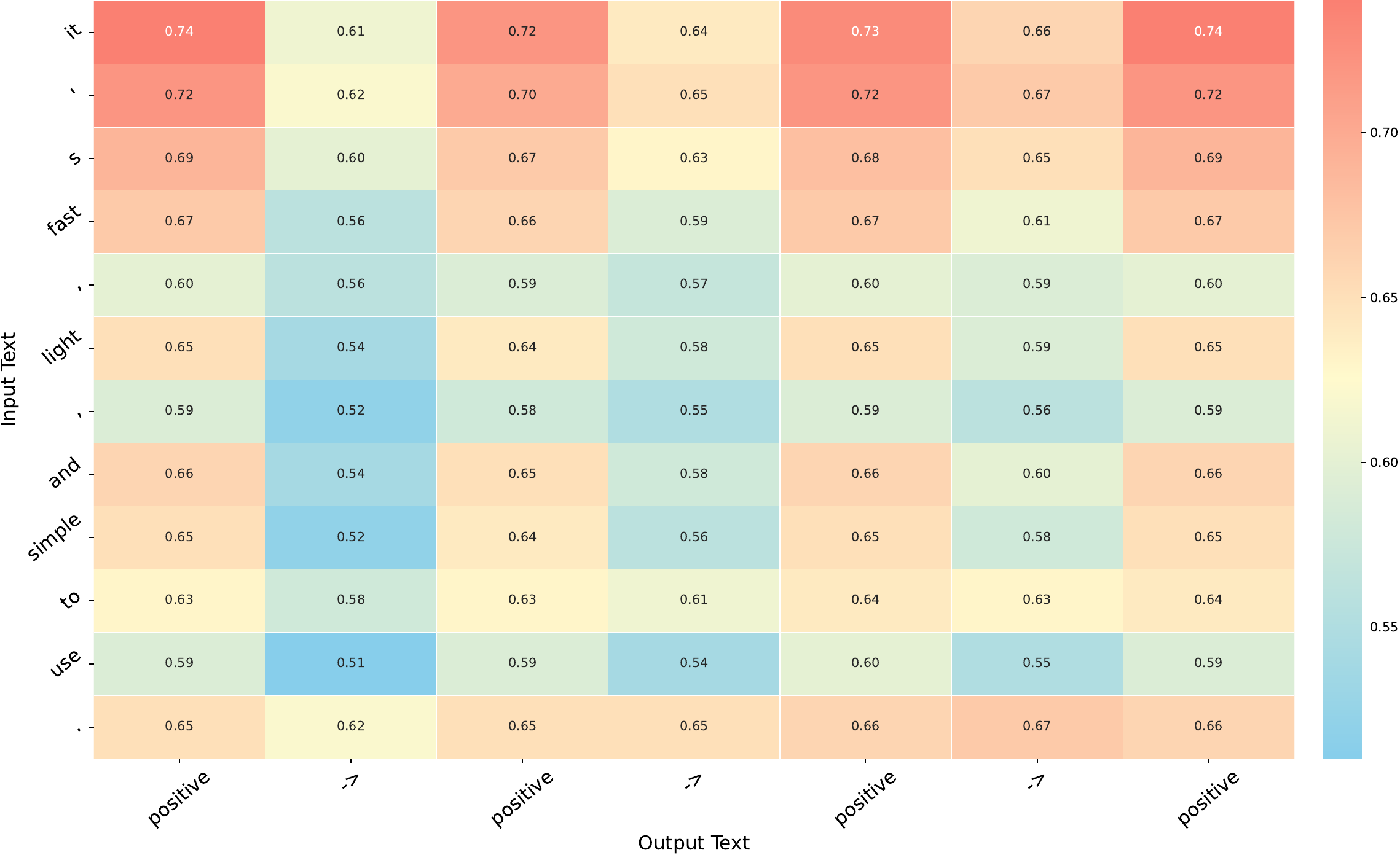}
    \\
    \includegraphics[width=0.25\textwidth]{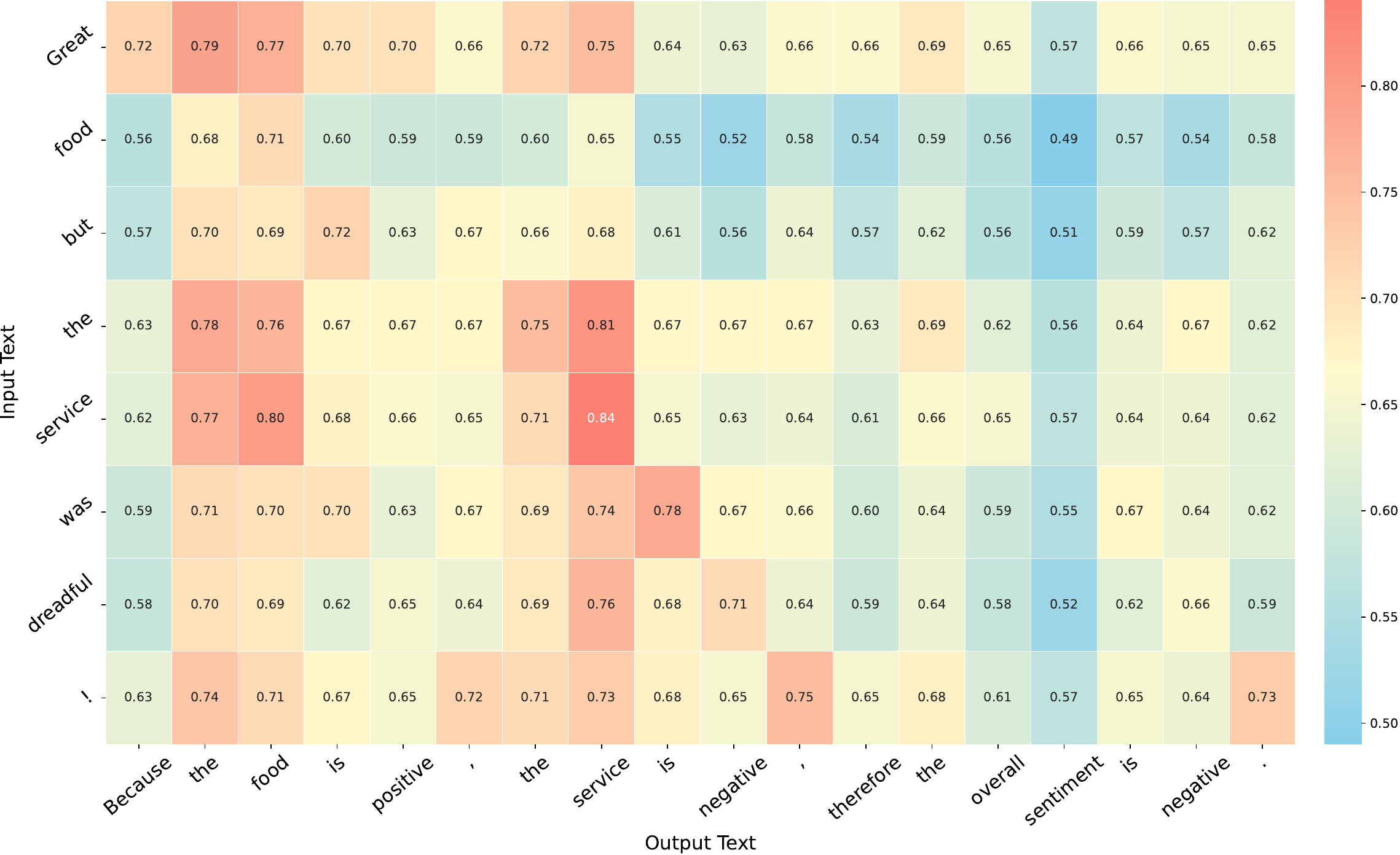}
    \includegraphics[width=0.25\textwidth]{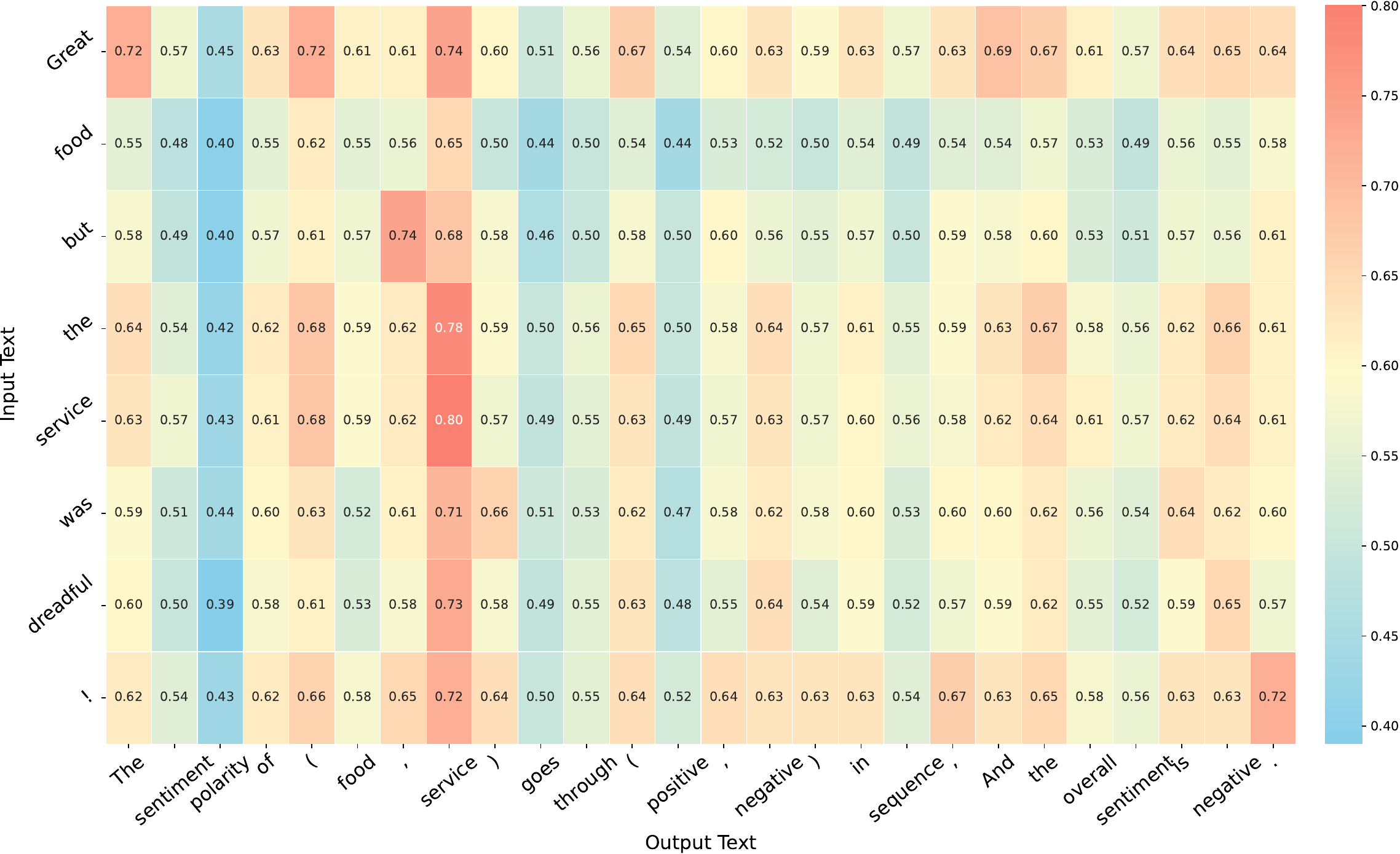}
    \includegraphics[width=0.25\textwidth]{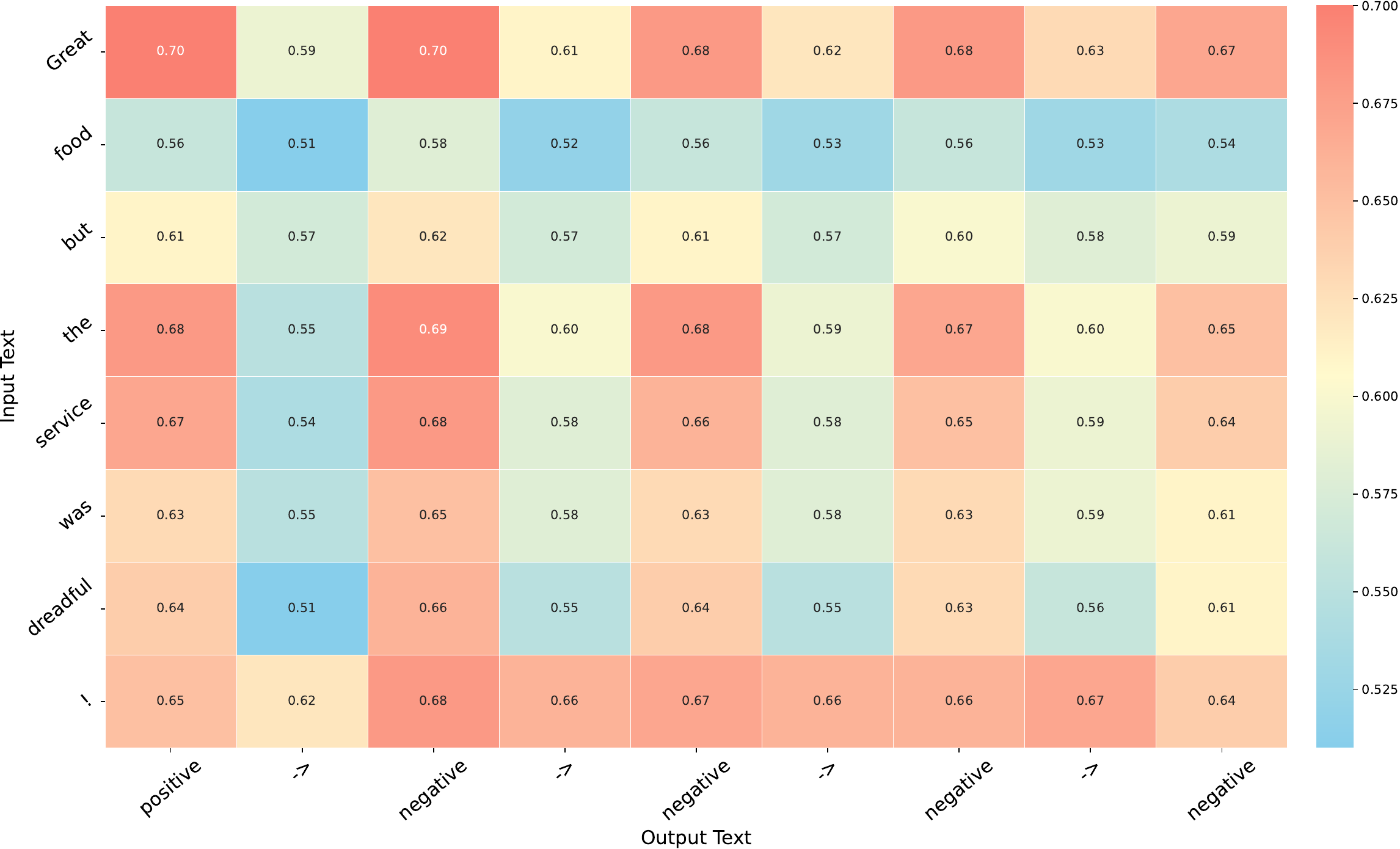}
    \end{tabular}
    \caption{Similarity between input and output tokens. Top row (Laptop) and bottom row (Restaurant) showing prompts for CoT-{\color{lightgreen} v1}, -{\color{lightgreen} v2}, and -{\color{lightgreen} v3} (Gemma2-2b with 18-shot on explicit split is reported).}
    \label{fig:question2_explicit}
\end{figure*}

\begin{figure*}[!ht]
    \vspace{-25pt}
    \centering
    \begin{tabular}{ccc}
    \includegraphics[width=0.25\textwidth]{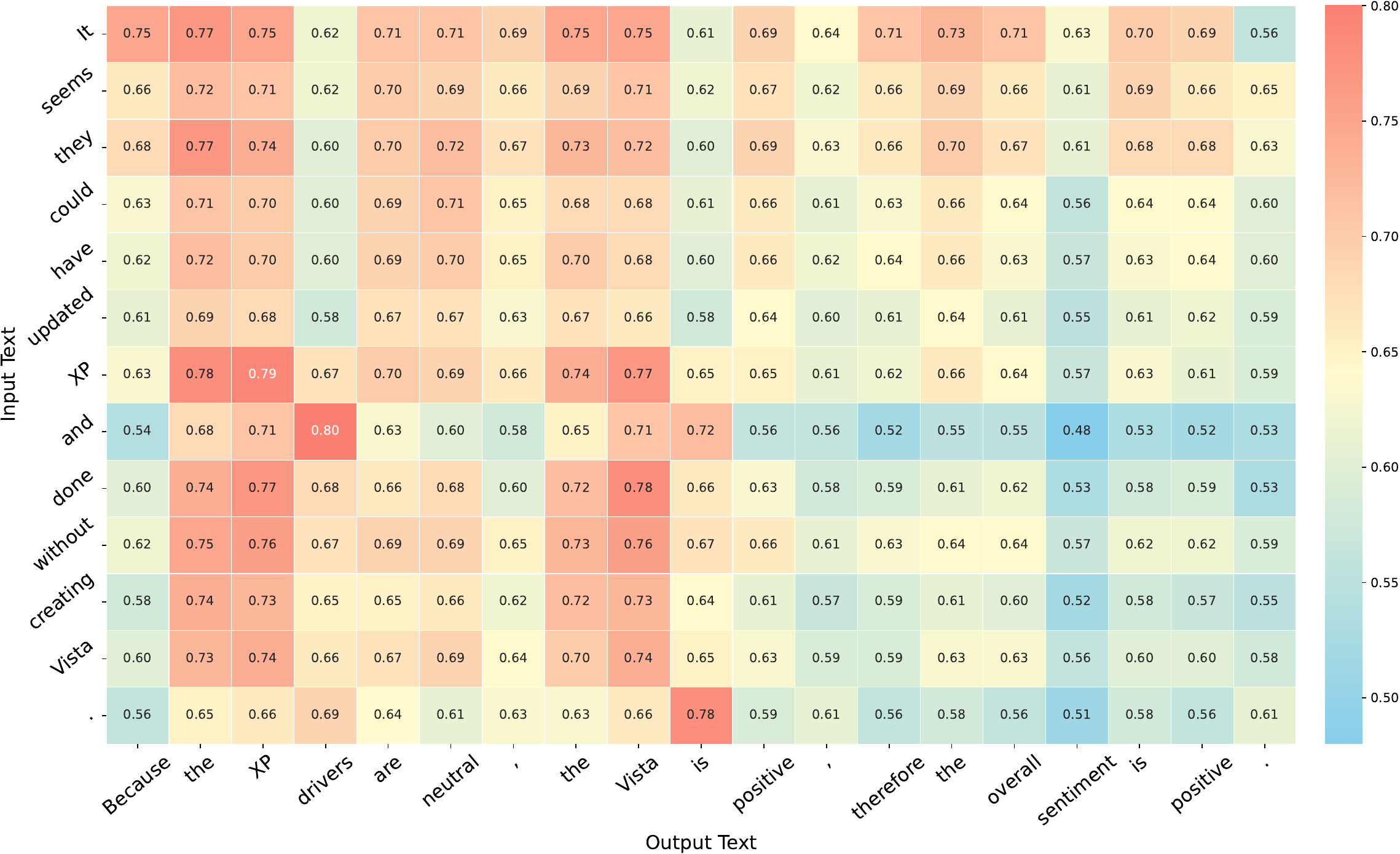}
    \includegraphics[width=0.25\textwidth]{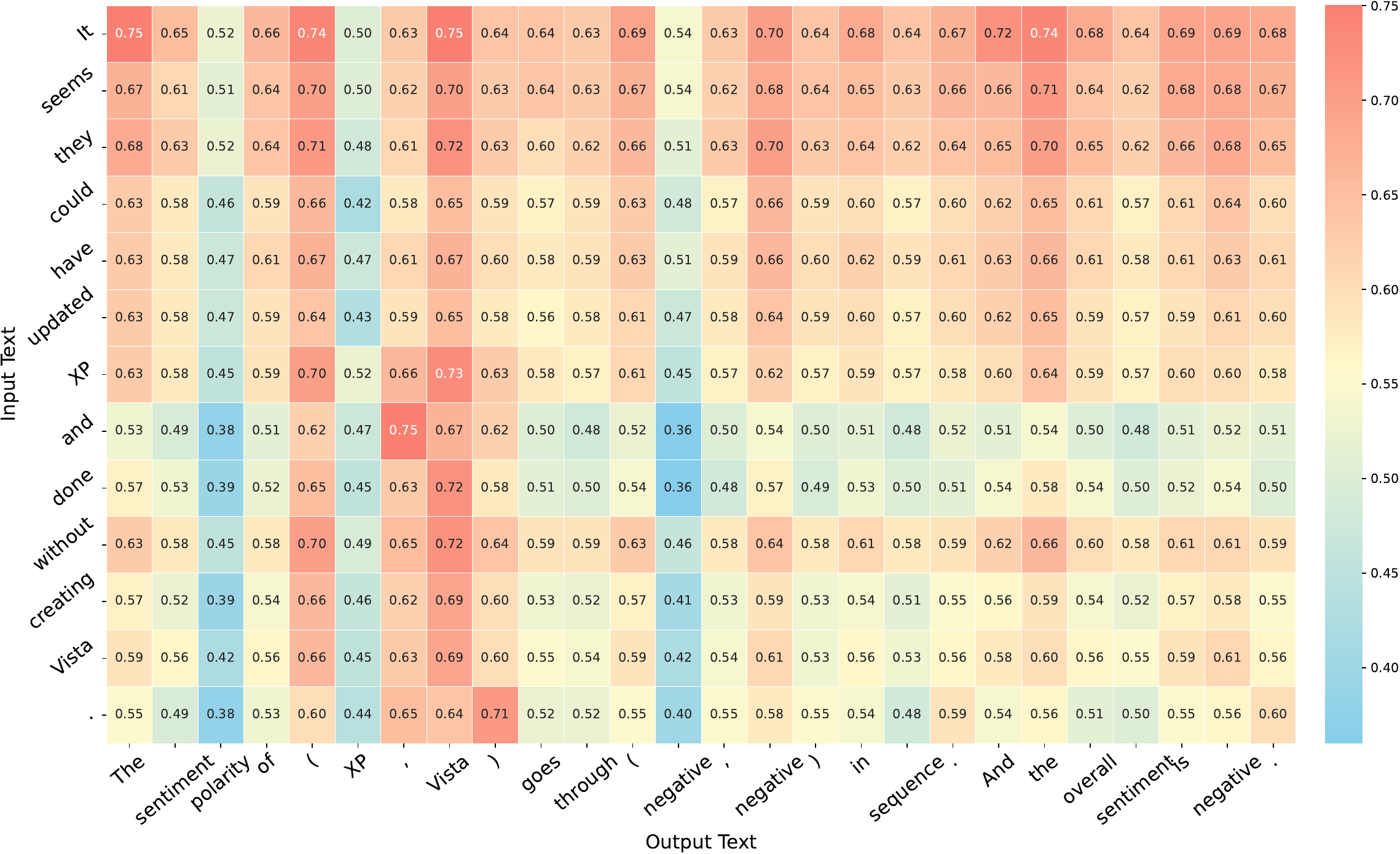}
    \includegraphics[width=0.25\textwidth]{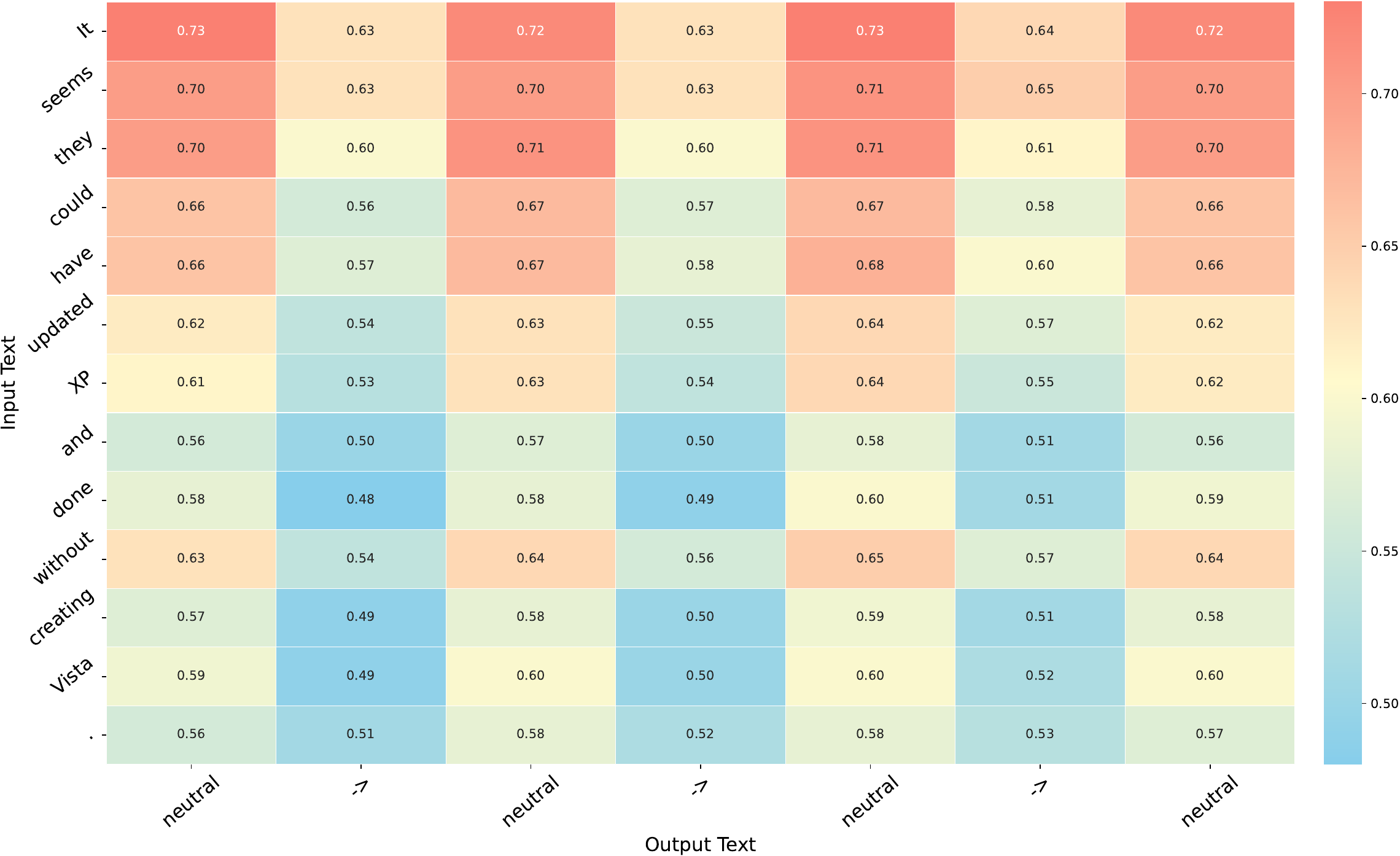}
    \\
    \includegraphics[width=0.25\textwidth]{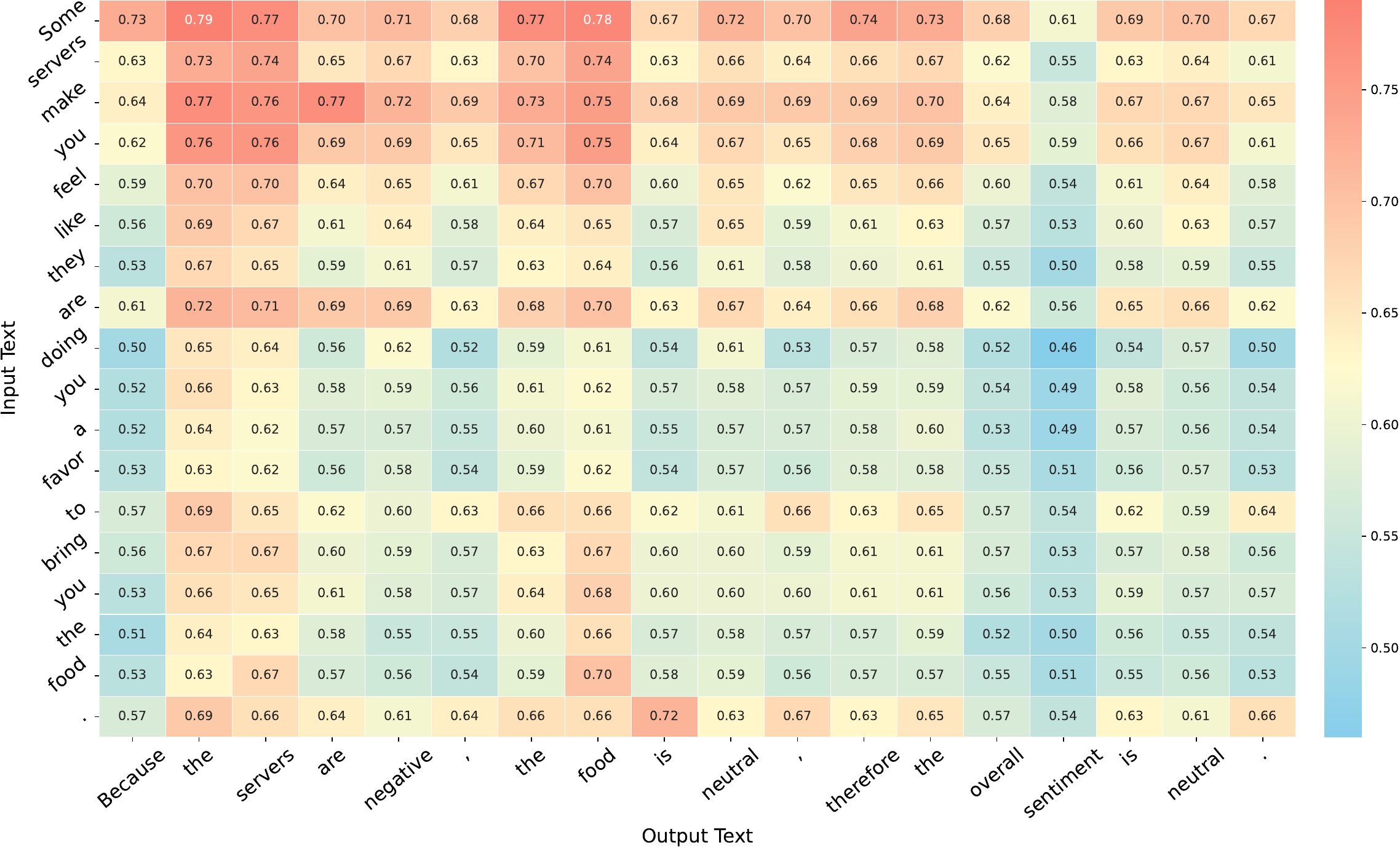}
    \includegraphics[width=0.25\textwidth]{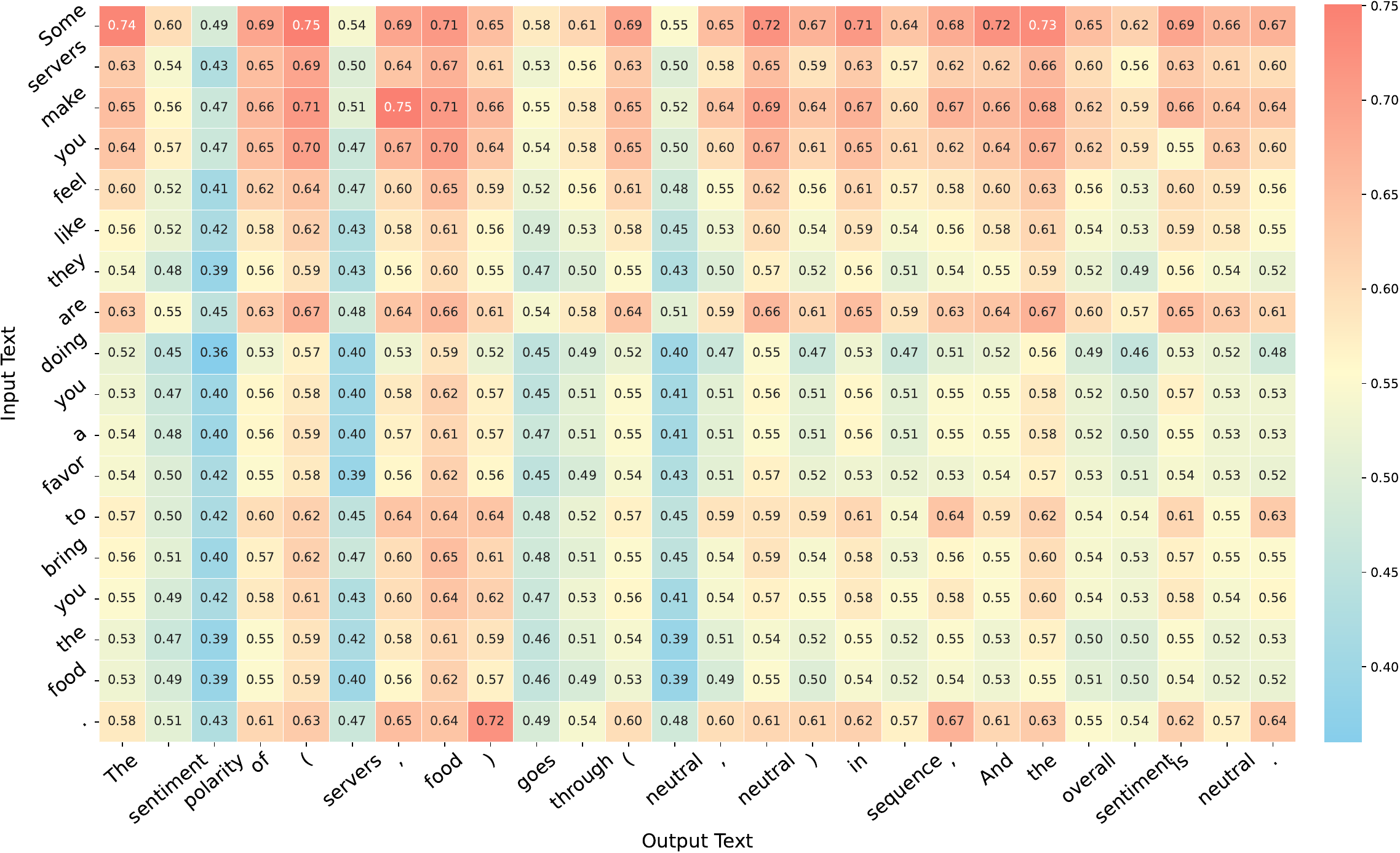}
    \includegraphics[width=0.25\textwidth]{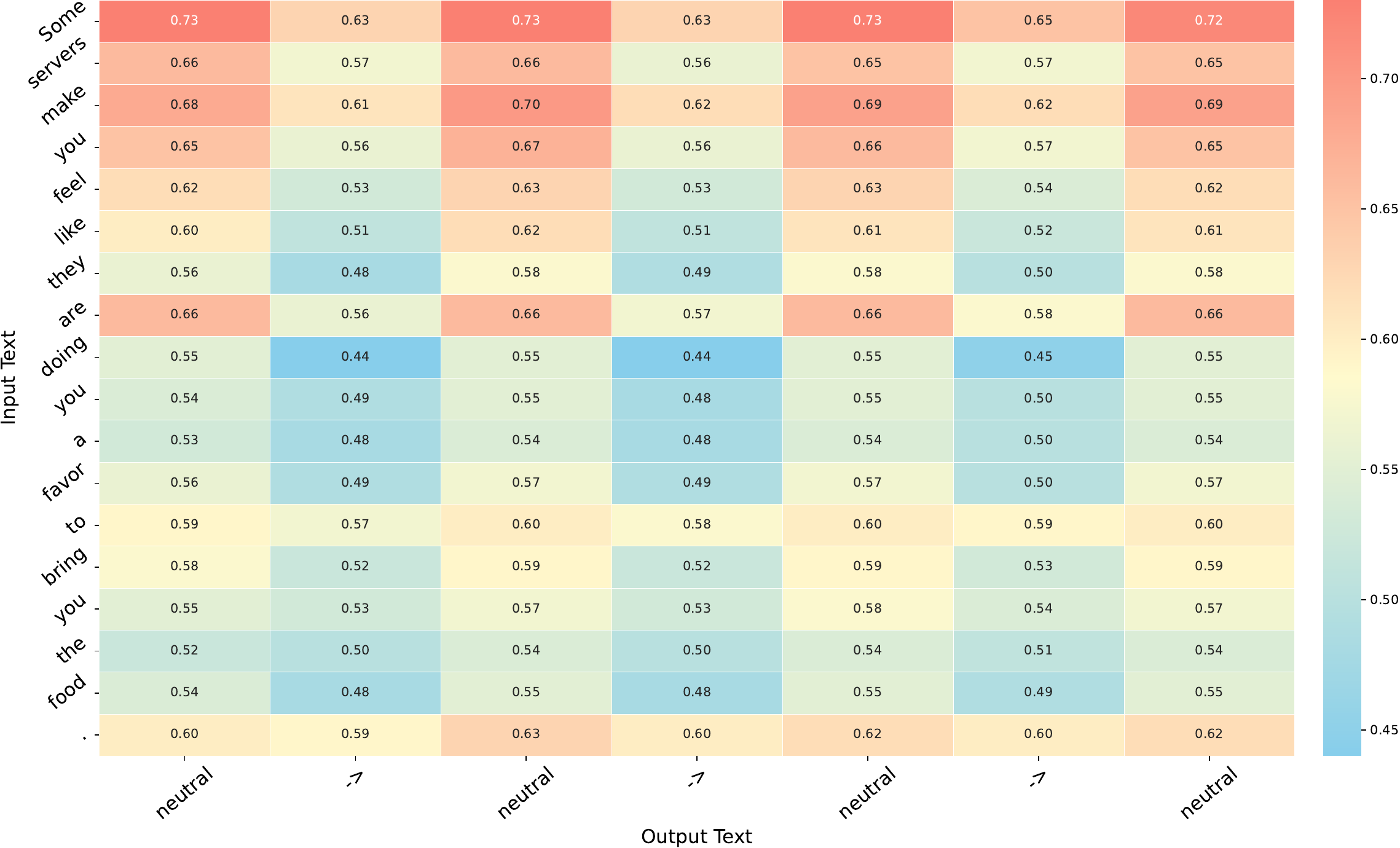}
    \end{tabular}
    \caption{Similarity between input and output tokens. Top row (Laptop) and bottom row (Restaurant) showing prompts for CoT-{\color{lightgreen} v1}, -{\color{lightgreen} v2}, and -{\color{lightgreen} v3} (Gemma2-2b with 18-shot on implicit split is reported).}
    \label{fig:question2_implicit}
    \vspace{-20pt}
\end{figure*}

Figures \ref{fig:question2_explicit} and \ref{fig:question2_implicit} illustrate our findings, which reveal several key insights: Firstly, the explicit split demonstrates a stronger similarity between sentiment words in the output and input text. In contrast, the implicit split does not exhibit such clear patterns. Moreover, the similarity between input text and aspect words is generally higher than the similarity between input text and corresponding aspect sentiments. Additionally, the standard prompt and various CoT versions maintained a generally consistent similarity between the overall sentiment of the output and the input text. This suggests that CoT may not have substantially influenced the model's interpretation of sentiment in the input text.

\subsection{RQ-4: Pre-training knowledge vs. demonstration information}

\begin{figure*}[!t]
    \centering
    \includegraphics[width=0.24\linewidth]{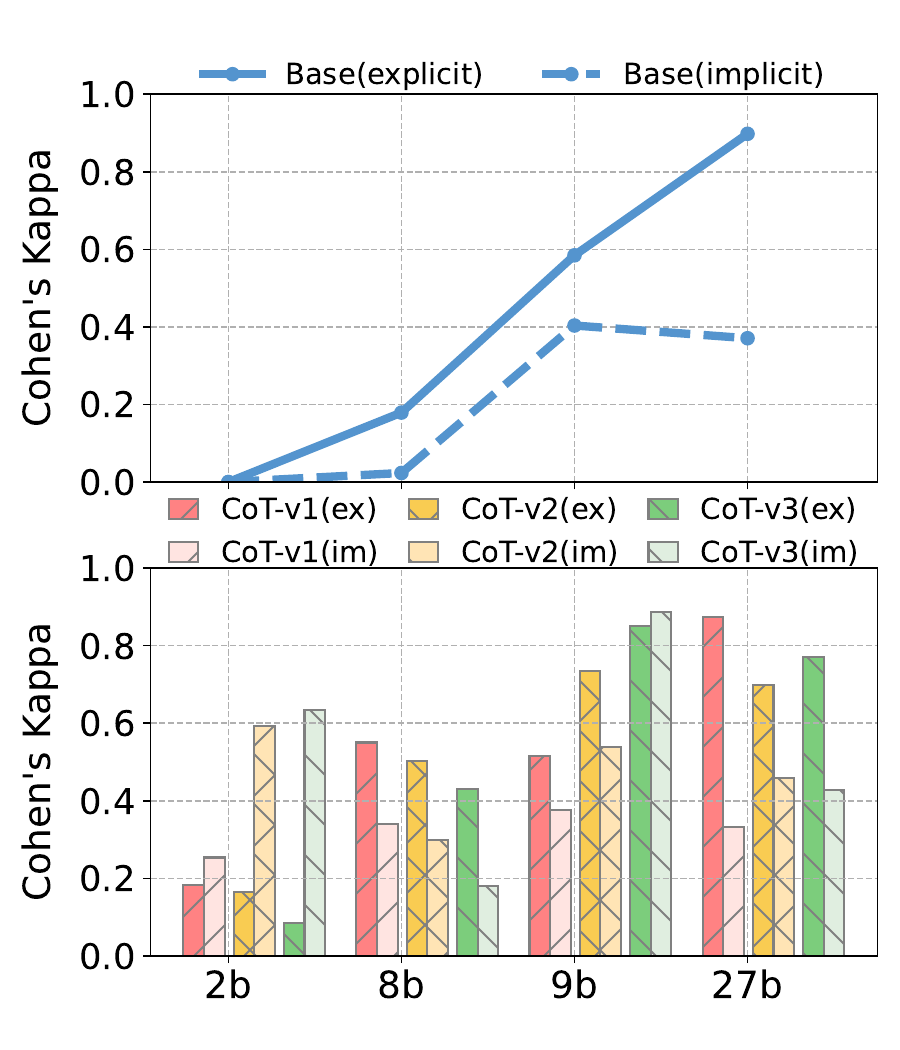}
    \includegraphics[width=0.24\linewidth]{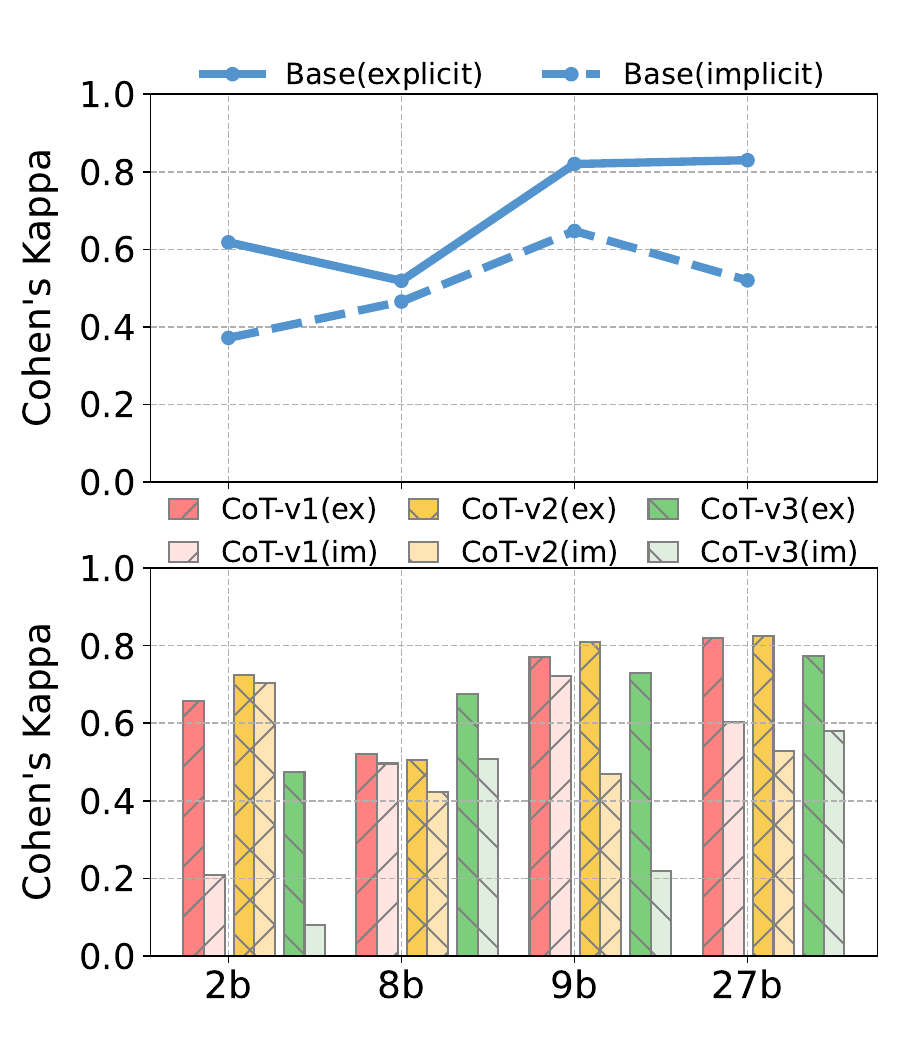}
    \includegraphics[width=0.24\linewidth]{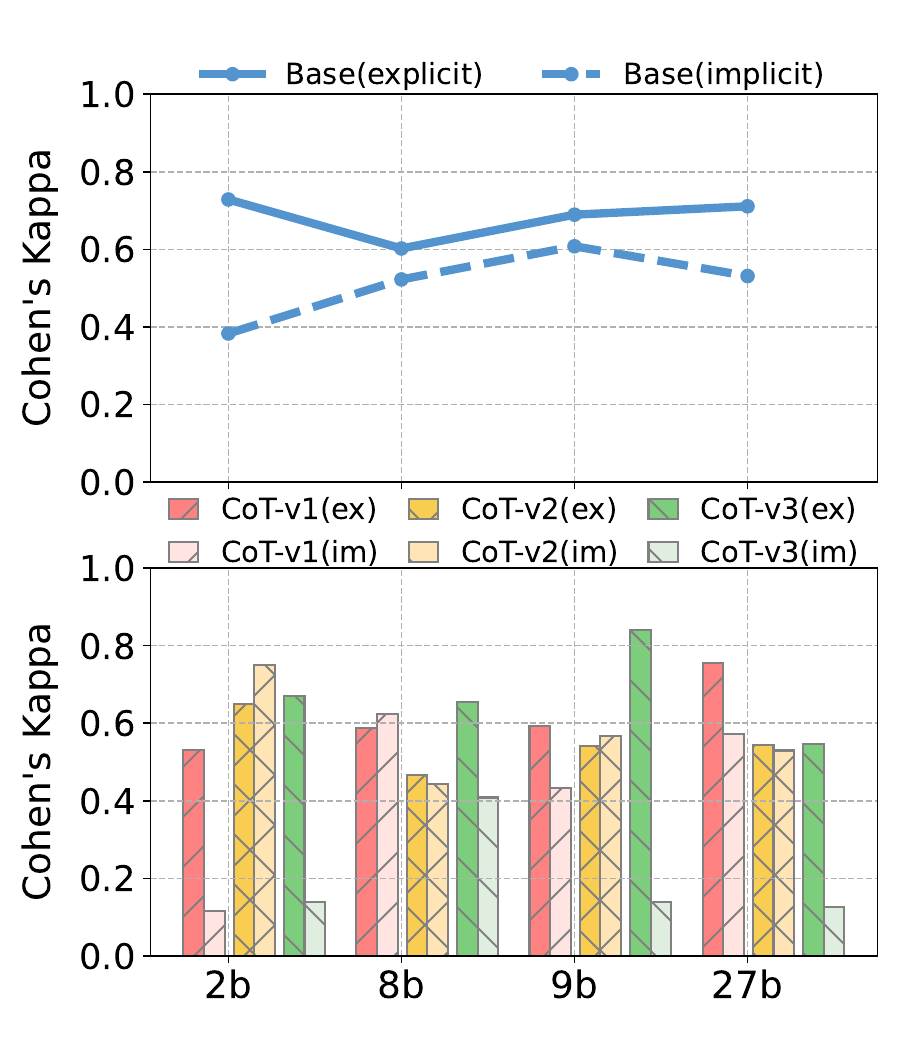}
    \includegraphics[width=0.24\linewidth]{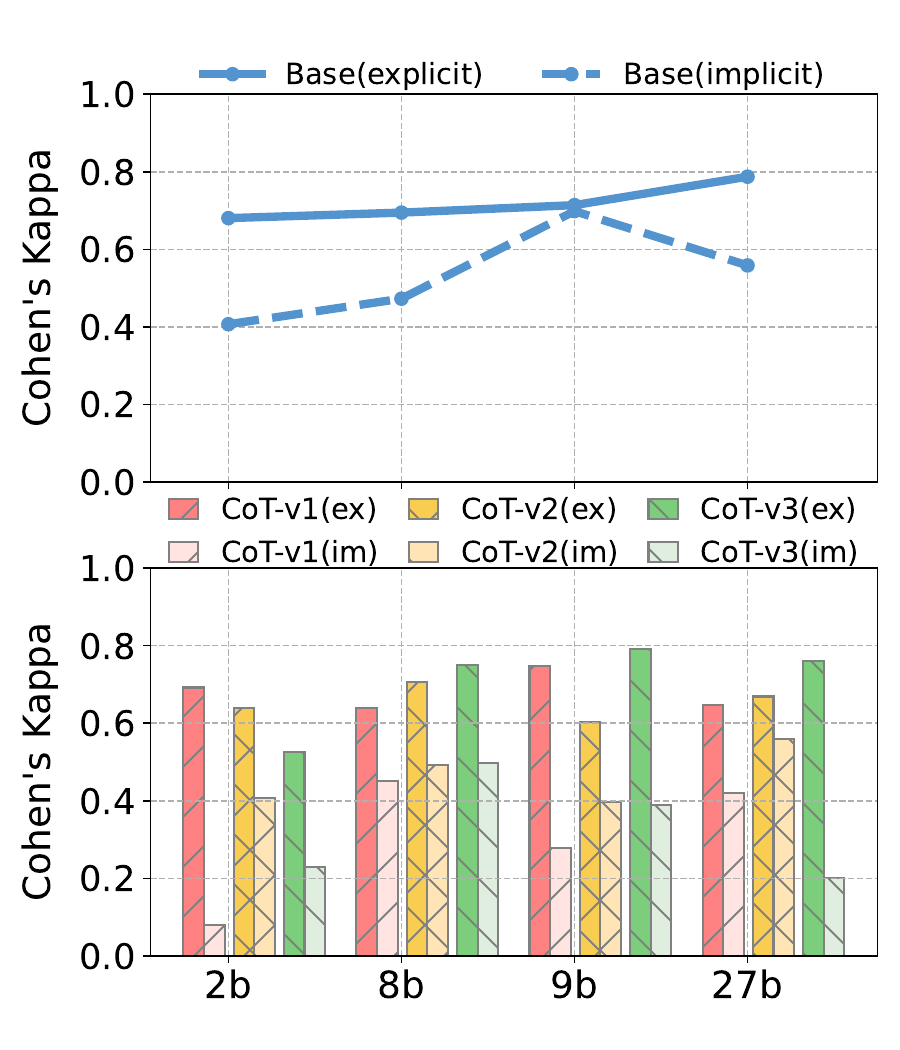}

    \includegraphics[width=0.24\linewidth]{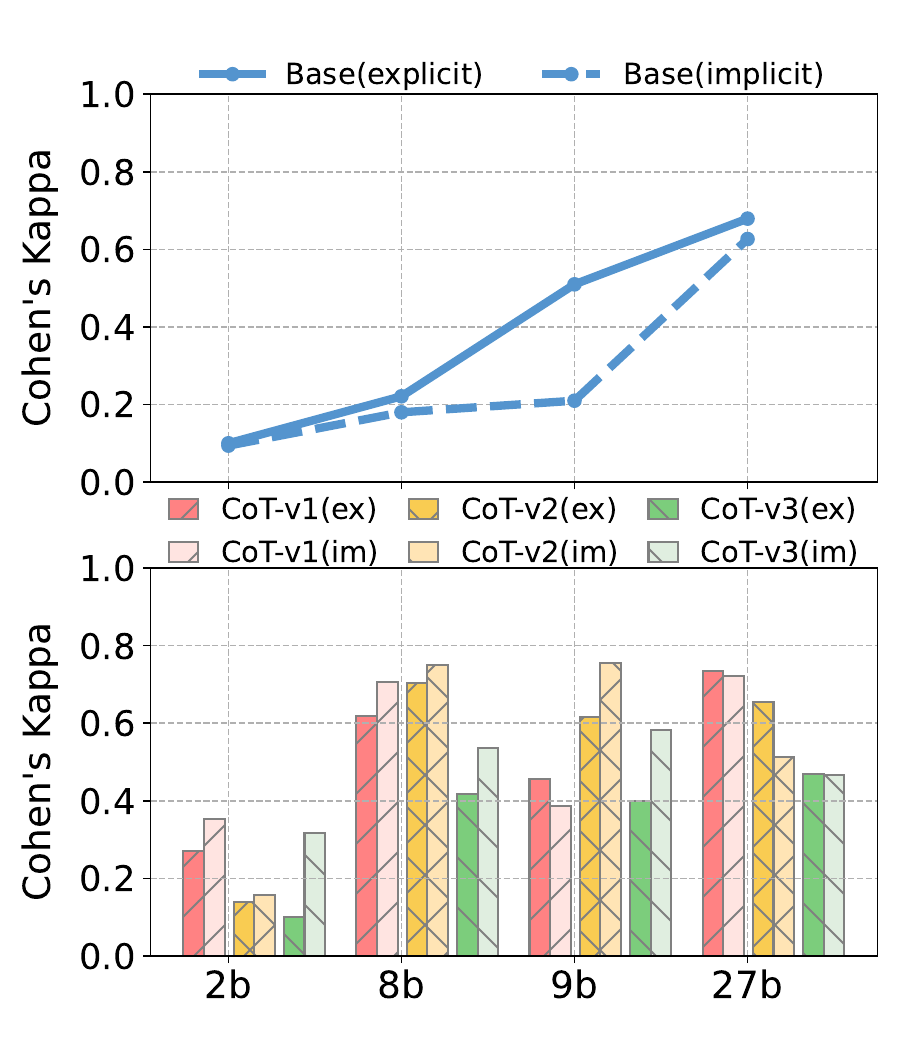}
    \includegraphics[width=0.24\linewidth]{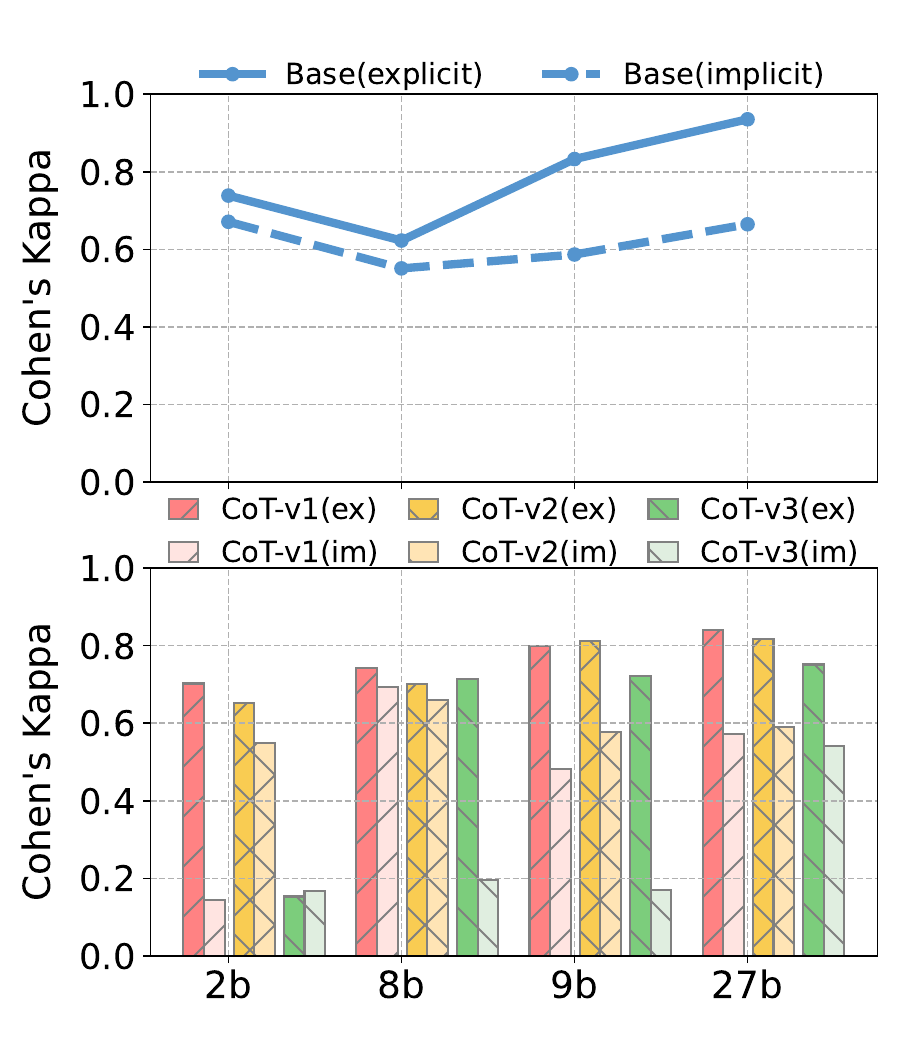}
    \includegraphics[width=0.24\linewidth]{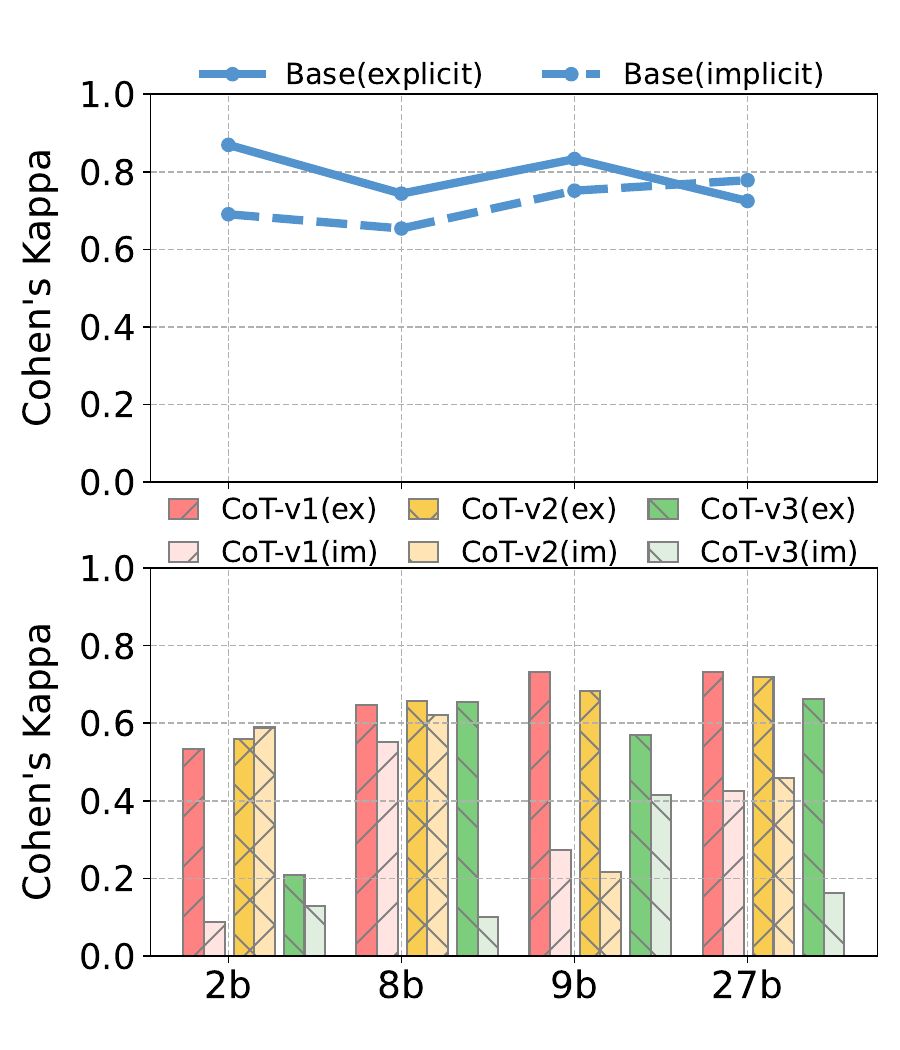}
    \includegraphics[width=0.24\linewidth]{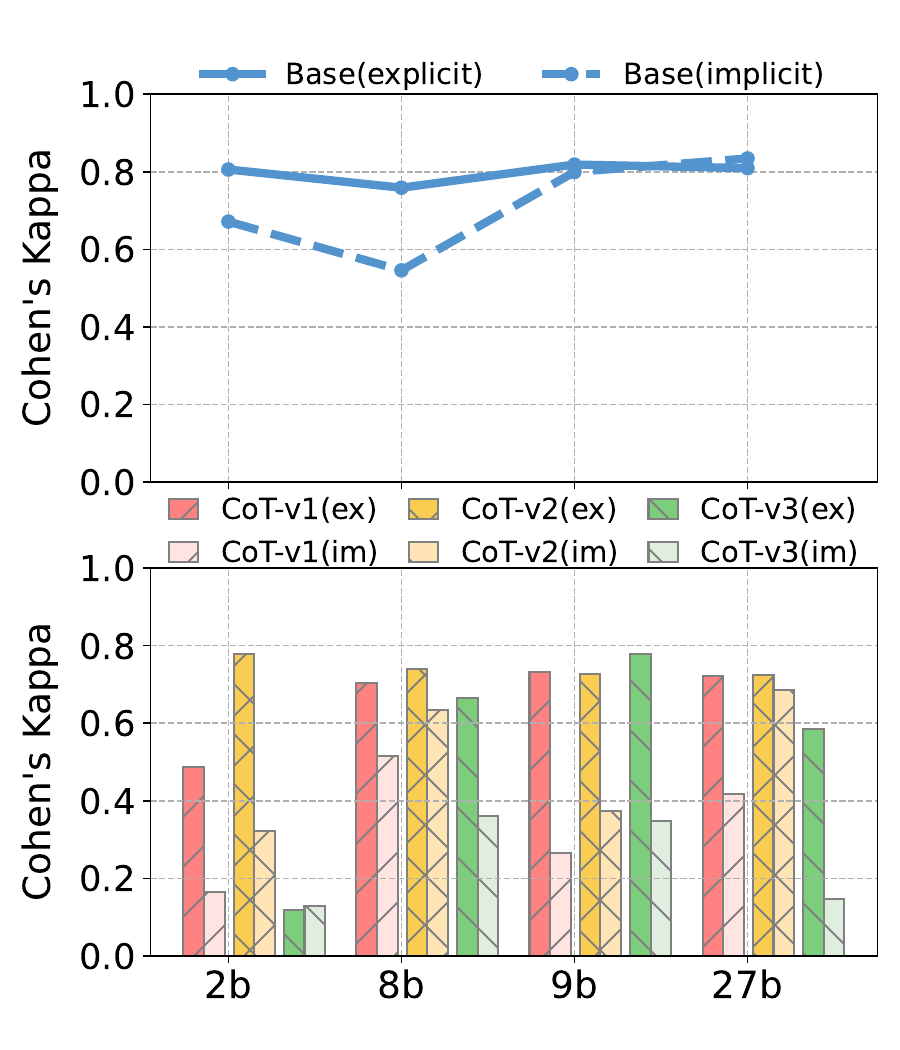}

    \caption{Model output agreement before and after input shuffling (upper part), and output agreement before and after sentiment reversal (lower part). Results for Laptop (top row) and Restaurant (bottom row). Due to space constraints, only the results for 1-, 4-, 8- and 12-shot scenarios are shown. } 
    \vspace{-15pt}
    \label{fig:question3}
\end{figure*}

Prior research on models like BERT has shown that word order shuffling in SA tasks has minimal impact on model performance~\cite{pham-etal-2021-order,sinha-etal-2021-masked}. Given that CoT is designed to reduce reasoning complexity through step-by-step processing, we investigate the extent to which the model's sentiment analysis relies on pre-training knowledge versus information provided in few-shot demonstrations, by examining whether word order disruption affects SA results undering CoT prompting.

\textbf{Word order disruption test.} We employed the word order disruption method from \cite{ZHAO2024122700}, sequentially swapping adjacent words to disrupt both local and global word positions. This process was applied only to the input question, leaving the demonstrations unchanged. To assess the impact of this disruption, we measured the agreement between predictions of disturbed and original inputs.

Results, as shown in the upper part of Figure \ref{fig:question3}. After perturbing the input text, model size positively correlates with agreement: Gemma-27b achieves higher mean agreement ({\color{orange} 0.79}) compared to the Gemma-2b ({\color{orange} 0.58}) on the explicit split of Laptop dataset. Moreover, the agreement strengthens with an increasing number of few-shot examples, as demonstrated by the Gemma-2b's performance on the explicit split of the Laptop dataset (1-shot: {\color{orange} 0.0}, 12-shot: {\color{orange} 0.70}). Additionally, explicit splits show higher mean agreement than implicit splits ({\color{orange} 0.67} vs. {\color{orange} 0.54}). Taking Gemma-27b as an exemplar, the model exhibited relatively small  variations in its generated content (i.e., higher agreement), with mean agreement values of {\color{orange} 0.80} and {\color{orange} 0.61} for explicit and implicit splits, respectively.

\textbf{Counterfactual demonstration test.} To further investigate the model's utilization of demonstration information, we adopted the counterfactual method proposed by \cite{madaan2023makes}. This approach involves creating a deliberate conflict between the knowledge in demonstrations and the presumed factual knowledge from the pre-training corpus.

We reversed the sentiment of aspects in the demonstrations, randomly replacing original sentiments with their opposites (positive with negative, negative with positive, and neutral with either positive or negative). The input questions remained unchanged. We also report the agreement between predictions of original and modified demonstrations. Results, presented in the lower part of Figure \ref{fig:question3}: When perturbing demonstrations, model size again correlates positively with agreement: Gemma-27b shows higher mean agreement ({\color{orange} 0.71}) compared to Gemma-2b ({\color{orange} 0.46}) on the explicit split of Laptop dataset. However, unlike input perturbation, increasing the number of few-shot examples leads to lower agreement, as evidenced by Gemma-27b's performance on the explicit split of Laptop dataset (4-shot: {\color{orange} 0.81}, 12-shot: {\color{orange} 0.69}). 

These results indicate that modifications in demonstrations significantly influenced the model's decisions, suggesting that the model  relies on demonstration information in SA tasks.


\section{Discussion of Language and Thought}
Our findings challenge Wittgenstein's view that ``language limits the boundaries of thought'' and support the independence of language and thought. However, the authors still align with Wittgenstein’s perspective.

Our results prompt deeper reflection: language may serve merely as a tool, whose role is to propagate and communicate abstract concepts, and these concepts must exist first before language can express them. Language is like a quantitative metric used to reflect the level of thinking ability.  From a static perspective, language cannot convey ideas beyond the scope of cognition. However, from a dynamic viewpoint, the progression of thought drives language evolution, while the expansion of language, in turn, facilitates deeper thinking. The following examples illustrate this perspective:

\textbf{Cultural differences shape language interpretation.} For instance, ``Nobody loves you'' carries a negative connotation in Western cultures, often inducing psychological distress \cite{Sapir1921}. Language can convey the concept of childbirth pain but cannot fully replicate the experience \cite{Scarry1985}. Similarly, the Yang-Mills equations, though containing complex formulas, require deep understanding to grasp their true meaning \cite{YangMills1954}. The emergence of new concepts like ``autonomous driving'' or ``Mars colonization'' has led to corresponding terms, expanding language boundaries \cite{Chomsky2002}. 

\textbf{Language development promotes Thought development.}  
The progress of language also promotes the development of thought. This is reflected in the communication function of language, where new thoughts are spread to others who do not possess them, thereby enabling those people to acquire the corresponding thinking ability through understanding language.

\section{Conclusion}

This paper refines the ``language and thought'' debate by framing language as sentiment understanding and thought as chain-of-thought. Experiments on two public datasets and one constructed emotional dataset show that chain-of-thought has limited impact on sentiment analysis, supporting Fedorenko's view on the independence of language and thought.



\bibliographystyle{splncs04}
\bibliography{custom}

\appendix

\section{Supplement}

\begin{table}[htbp]
    \centering
    \setlength{\extrarowheight}{2pt}
    \caption{CoT Prompts Examples}
    \begin{tabular}{|p{0.18\textwidth}|p{0.82\textwidth}|}
    \hline
    \rowcolor{gray!10}
    Input & Battery life could be better but overall for the price and Toshiba's reputation for laptops it's great! \\
    \hline\hline
    \textbf{Type} & \textbf{Content} \\
    \hline
    CoT-v1 & Because the Battery life is negative, the price is positive, therefore the overall sentiment is positive. \\
    \hline
    CoT-v2 & The sentiment polarity of (Battery life, price) goes through (negative, positive) in sequence. And the overall sentiment is positive. \\
    \hline
    CoT-v3 & negative -> positive -> positive \\
    \hline
    \end{tabular}
    \label{tab:cot-examples}
\end{table}

\end{document}